\documentclass[10pt,twocolumn,letterpaper]{article}

\usepackage{cvpr}              % To produce the CAMERA-READY version
\usepackage{tcolorbox}
\usepackage{listings}  % For displaying JSON or code
\usepackage{xcolor}    % For color customization
\tcbuselibrary{listingsutf8,breakable}
\usepackage{subcaption}
\usepackage{multicol}

\definecolor{cvprblue}{rgb}{0.21,0.49,0.74}
\usepackage[pagebackref,breaklinks,colorlinks,allcolors=cvprblue]{hyperref}

\usepackage{url}

\usepackage{algorithm}
\usepackage{algorithmic}
\usepackage{booktabs}
\usepackage{times}
\usepackage{epsfig}
\usepackage{amsmath}
\usepackage{amssymb}
\usepackage{caption}
\usepackage{multirow}
\usepackage[flushleft]{threeparttable}
\usepackage{pifont}% http://ctan.org/pkg/pifont

\usepackage{color}
\usepackage{array}
\usepackage{makecell}

\newlength\savewidth
\usepackage[utf8]{inputenc} % allow utf-8 input
\usepackage[T1]{fontenc}    % use 8-bit T1 fonts
\usepackage{hyperref}       % hyperlinks
\usepackage{amsfonts}       % blackboard math symbols
\usepackage{nicefrac}       % compact symbols for 1/2, etc.
\usepackage{microtype}      % microtypography
\usepackage{xcolor}         % colors

\usepackage{tabularray}
\usepackage{colortbl}
\usepackage{arydshln}

\definecolor{mygray}{gray}{.9}
\definecolor{mygreen}{rgb}{0, 0.6, 0}
\definecolor{myteal}{rgb}{0.16, 0.47, 0.56}
\definecolor{mypink}{rgb}{0.81, 0.25, 0.44}

\def\paperID{8} % *** Enter the Paper ID here
\def\confName{CVPR}
\def\confYear{2025}

\title{
A Large-Scale Analysis on Contextual Self-Supervised Video Representation Learning
}

\author{Akash Kumar$^{1 \dagger} $  \qquad Ashlesha Kumar$^{2}$  \qquad Vibhav Vineet$^{3}$ \qquad  Yogesh Singh Rawat$^{1}$\\
CRCV, University of Central Florida$^{1}$ \qquad BITS Pilani$^{2}$ \qquad Microsoft Research$^{3}$\\
}

\begin{document}

\maketitle

% \iffalse
\begin{abstract}

Self-supervised learning has emerged as a powerful paradigm for label-free model pretraining, particularly in the video domain, where manual annotation is costly and time-intensive. However, existing self-supervised approaches employ diverse experimental setups, making direct comparisons challenging due to the absence of a standardized benchmark. In this work, we establish a unified benchmark that enables fair comparisons across different methods. Additionally, we systematically investigate five critical aspects of self-supervised learning in videos: (1) dataset size, (2) model complexity, (3) data distribution, (4) data noise, and (5) feature representations. To facilitate this study, we evaluate six self-supervised learning methods across six network architectures, conducting extensive experiments on five benchmark datasets and assessing performance on two distinct downstream tasks. Our analysis reveals key insights into the interplay between pretraining strategies, dataset characteristics, pretext tasks, and model architectures. Furthermore, we extend these findings to Video Foundation Models (ViFMs), demonstrating their relevance in large-scale video representation learning. Finally, leveraging these insights, we propose a novel approach that significantly reduces training data requirements while surpassing state-of-the-art methods that rely on 10× more pretraining data. We believe this work will guide future research toward a deeper understanding of self-supervised video representation learning and its broader implications.
\end{abstract}

\section{Introduction}
\label{sec:intro}

\footnote{$^{\dagger}$Corresponding Author: akash.kumar@ucf.edu}

Deep learning models require a large amount of labeled data for their training. Obtaining annotations at large-scale needs a lot of effort and it becomes even more challenging as we shift from image to video domain.  
There are several interesting directions focusing on this issue such as domain adaptation \cite{da}, knowledge distillation \cite{kd}, semi-supervised learning \cite{semi}, self-supervision \cite{main_survey} and weakly-supervised learning \cite{weakly}, which attempts to rely on the knowledge learned from existing source datasets and transfer to new target datasets with minimal labels.  Among these approaches, self-supervised learning use pretext task as supervisory signal and does not require any labels on source datasets  which makes it more favorable.

 \begin{figure*}[t!]
     \centering
   \includegraphics[width=\linewidth]{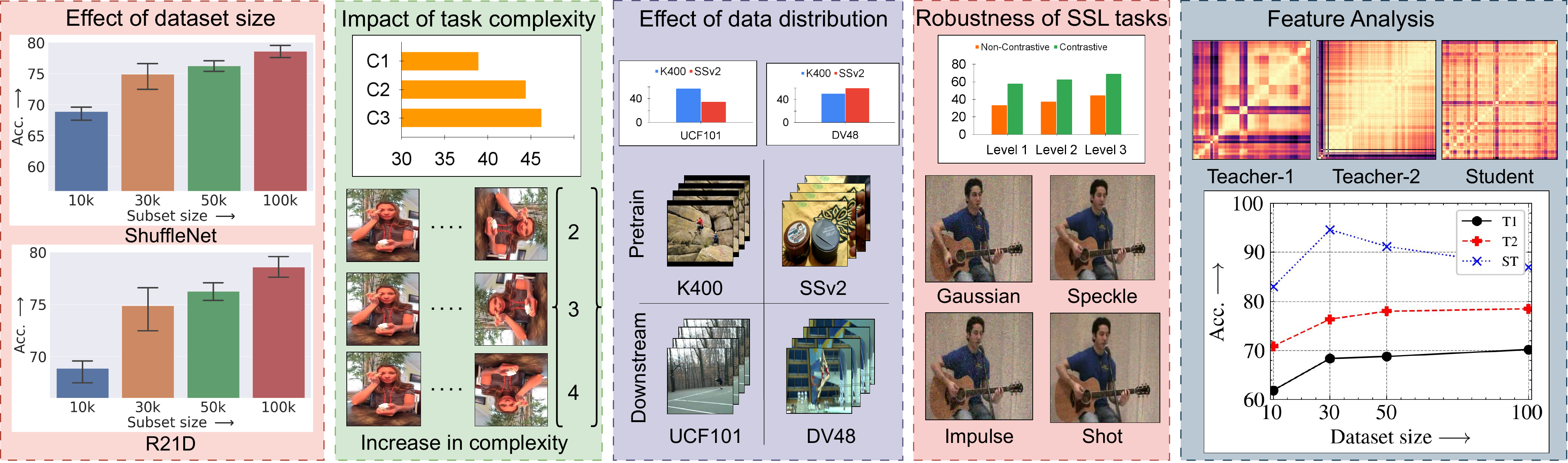}
     \caption{\textbf{Overview of proposed benchmark.} We study five different aspects in this benchmark study.
     Starting from left, 1) we show the analysis of \textit{effect of dataset size vs training time}. As the dataset size increases, variation in performance decreases even with longer training time, 2) We show the effect of \textit{task complexity} (C1, C2, C3 - Different complexities). Bottom figure shows use case of how complexity increases for the RotNet task, and, top figure shows how the performance varies for the R21D network, 3) With different \textit{data distribution shifts}, the third sub-figure shows the impact of \textit{target} data distribution on the \textit{source} data, 4) We look into another data distribution shift due to introduction of noise. We see how \textit{non-contrastive }tasks are more robust than \textit{contrastive} ones even with increasing levels of severity of noise. The bottom part shows an example for each type of noise. Clips are provided in supplementary, and, 5) Finally, we further analyze whether the features learn \textit{orthogonal} information. In this sub-figure, we show that using different architectures as teachers can substantially improve performance even in a low-data regime.
     }
\label{fig:main_fig}
\vspace{-15pt}
 \end{figure*}

In recent years, self-supervised learning (SSL) has made significant progress in video representation learning \cite{vcop, rotnet, prp, tdl, cvrl, rspnet}. More recently, research has shifted towards context-based learning, which involves modifying input data to derive a classification \cite{pace, tclr, vcop, rotnet}, reconstruction \cite{prp, rspnet}, or generative \cite{gen_1, gen_2, mem_dpc, videomae, videomoco} signal that serves as a learning objective. The primary focus of these approaches is to design pretext tasks that are computationally efficient while providing a strong supervisory signal, enabling models to learn meaningful \textit{spatio-temporal} features.  

Despite these advancements, comparing different SSL methods remains challenging due to the absence of standardized evaluation protocols. Existing approaches are assessed under varying conditions, lacking a common benchmark to ensure fair and consistent evaluation. A recent study \cite{Thoker2022HowSI} takes a step in this direction by analyzing downstream learning; however, it does not explore the self-supervision aspect, which is the primary focus of our work.  

In this study, we introduce a benchmark where key self-supervised pretraining parameters are standardized across methods to enable fair comparisons. Using this benchmark, we systematically investigate five critical aspects of self-supervised learning: (1) the effect of pretraining dataset size, (2) task complexity, (3) generalization under distribution shifts, (4) robustness against data noise, and (5) properties of the learned representations. Fig. \ref{fig:main_fig} provides an overview.  

Our benchmark conducts a large-scale evaluation of representative context-based self-supervised methods for video representation learning. We analyze two key factors: (1) the \textit{learning objective}, differentiating between \textit{contrastive} and \textit{non-contrastive} methods, and (2) \textit{data transformations}, categorized into \textit{spatial}, \textit{temporal}, and \textit{spatio-temporal} variations. We examine six pretext tasks across six different architectures, conducting experiments on five action recognition datasets and evaluating these approaches on two downstream tasks: action recognition and video retrieval. Furthermore, we extend our study to recently developed video foundation models.  

Our findings reveal several key insights: (1) contrastive learning methods converge faster but exhibit lower robustness to data noise, (2) increasing pretext task complexity does not necessarily lead to better spatio-temporal representation learning, (3) \textit{temporal} pretext tasks are more challenging than \textit{spatial} or \textit{spatio-temporal} ones, (4) spatio-temporal tasks demonstrate robustness to distribution shifts, and (5) pretext tasks learn complementary features across different architectures, dataset distributions, dataset sizes, and pretext task types.  

Our contributions are threefold:  
\begin{itemize}  
    \item We introduce a benchmark for self-supervised video representation learning, ensuring fair comparisons of different pretext tasks under a unified experimental setup.  
    \item We conduct an extensive analysis of five critical factors in self-supervised learning for videos: (1) dataset size, (2) task complexity, (3) distribution shifts, (4) data noise, and (5) feature representations.  
    \item Finally, we validate our insights by proposing a simple yet effective approach that surpasses existing state-of-the-art methods on video action recognition while requiring significantly less pretraining data. Additionally, we provide a structured recipe for future self-supervised learning methods to build upon.  
\end{itemize}  

We believe our study will serve as a foundation for advancing self-supervised video representation learning and guiding future research in the field.

\section{Related Work}
\label{sec:rw}

\paragraph{Self-supervised learning}  
Several studies have explored self-supervised learning (SSL) for video representation learning \cite{main_survey, schiappa2022self} for efficient labeling amongst other approaches such as semi-supervised \cite{Kumar_2022_CVPR, modi2022video, Rana_2023_CVPR, Dave_2022_WACV, ayushneurips22, Singh_Rana_Kumar_Vyas_Rawat_2024, kumar2024stable} and weakly-supervised \cite{kumar2025contextual, garg2025stpro}. These approaches can be broadly categorized based on the nature of the pretext task: (1) context-based methods \cite{st_puzzles, context_2, vid_jigsaw, odd_one_out, pace, iic, visual_tempo, tclr, temporals, tdl, cvrl, rspnet, transrank, Guo2022CrossArchitectureSV, Ranasinghe2021SelfsupervisedVT} and (2) cross-modal methods \cite{cross_modal_1, cross_modal_2, cross_modal_3}.  Cross-modal approaches leverage multiple modalities, such as audio, video, optical flow, and camera viewpoints, relying on cross-modal consistency to learn representations. In contrast, context-based learning utilizes transformations within a single modality to generate supervisory signals. Over the years, context-based pretraining tasks have evolved significantly, focusing on understanding how different transformations impact learned representations. Unlike cross-modal approaches, context-based methods explicitly exploit spatial and temporal information through various transformations \cite{shuffle_learn, odd_one_out, vcop, speednet, pace, cvrl, tdl}. Recent works have extended these methods by jointly transforming both spatial and temporal domains \cite{st_puzzles, vcp, iic, prp, rspnet}. While incorporating multiple modalities can enhance performance, such data is often unavailable for large-scale datasets. Therefore, in this work, we focus exclusively on single-modality (RGB) approaches.  

\paragraph{Self-supervised benchmarking}  
Several prior studies have focused on benchmarking SSL in the image domain. In \cite{goyal2019scaling}, the authors provide an in-depth analysis of self-supervised image representation learning, investigating how dataset scaling influences learned representations. Similarly, \cite{revisit} examines the role of different model architectures in self-supervised learning. However, both studies primarily focus on downstream performance rather than analyzing the pretext tasks themselves. Their primary objective is to improve specific pretext tasks rather than to study their underlying impact on representation learning.  In contrast, we systematically analyze various pretext tasks and evaluate how different factors influence feature learning. Additionally, while previous studies primarily focus on the image domain, our work extends this analysis to videos.  \cite{unsupervised} explores unsupervised learning in the video domain by adapting pretext tasks from images to videos. However, their primary emphasis remains on downstream evaluation.  Our work differs by focusing on the self-supervised aspect itself, analyzing key factors such as dataset size, task complexity, data distribution, and robustness to noise. By addressing these aspects, we provide a comprehensive understanding of how different pretraining strategies affect video representation learning. A recent study \cite{kumar2023benchmarking} looks into similar aspect but hasn't looked into video foundation models which corresponds to the fture ahead.

\section{Self-Supervised Configurations}
We first describe the pretext tasks used in our study along with their categorization followed by details of this benchmark including
network architectures, datasets, downstream tasks and evaluations.

 \subsection{Task Categorization}  

We analyze video pretext tasks from two key perspectives: (1) the transformations applied to the data and (2) the learning objectives. The data transformations fall into three categories: \textit{spatial-based (S)}, \textit{temporal-based (T)}, and \textit{spatio-temporal (ST)}. \textit{Spatial} transformations involve reshuffling spatial patches, applying temporally consistent data augmentations, or rotating images and patches. \textit{Temporal} tasks focus on frame/clip permutation classification, order verification, clip sampling at different playback rates, or contrastive learning using temporal triplets. \textit{Spatio-temporal} tasks simultaneously modify both spatial and temporal dimensions. Examples include dilated sampling with frame reconstruction, joint spatial and temporal shuffling, speed prediction, and contrastive learning of visual features.  

The learning objectives can be categorized as either \textit{contrastive} \cite{Chen2020ASF} or \textit{non-contrastive} \cite{prp}. Based on this categorization, we select at least two representative pretext tasks from each transformation category, ensuring inclusion of both contrastive and non-contrastive methods. Specifically, we study the following pretext tasks: RotNet (Rot) \cite{rotnet}, Video Clip Order Prediction (VCOP) \cite{vcop}, Playback Rate Prediction (PRP) \cite{prp}, Spatiotemporal Contrastive Video Representation Learning (CVRL) \cite{cvrl}, Temporal Discriminative Learning (TDL) \cite{tdl}, and Relative Speed Perception Network (RSPNet) \cite{rspnet}. A detailed description of these tasks is provided in the supplementary material.
 
\subsection{Benchmark details}
\label{sec:bench_detail}
This section standardizes the conditions used by our benchmark to compare different pretext tasks. Further explanation for using these conditions are outlined in the supplementary.

\noindent{\textit{\textbf{Datasets:}}} 
We experiment with two different dataset types, 1) where appearance is more important, and 2) where time is more important. For appearance based, we use 
Kinetics-400 \cite{kinetics}, UCF101 \cite{ucf101}, and HMDB51 \cite{hmdb51}, where appearance is more important (recognize activity with a single frame) than temporal aspect, and for temporal aspect, we use Something Something-V2 \cite{ssv2} and Diving48 \cite{dv48}, where temporal information plays a significant role (require few frames to recognize activity). More details are in the supplementary.

\noindent{\textit{\textbf{Spatio-temporal architectures:}}}
We consider three different network capacities, 1) small-capacity, 2) medium-capacity, and large-capacity. For small capacity networks, we use ShuffleNet V1 2.0X \cite{shufflev1}, whereas for medium capacity we focus on R(2+1)D \cite{r21d} (R21D). We do not include large capacity networks in our main benchmark in the interest of computational efficiency; additional results for such a model, VideoSwin \cite{videoswin} is shown in the supplementary.

\noindent{\textit{\textbf{Downstream tasks:}}} We show results and analysis on two different downstream tasks - \textit{action recognition} and \textit{clip retrieval}. These two tasks are the most prominent in the field of self-supervised learning in videos. Full finetuning is performed as opposed to linear probing to adapt models. % to these downstream tasks.

\noindent{\textit{\textbf{Evaluation and Analysis:}}}We use top-1 accuracy for action recognition and top-K for Clip retrieval. For robustness performance, we calculate the relative robustness score $(R_{s})$ using original accuracy on clean test set $(A_{c})$ and perturbed accuracy on noisy test set$(A_{p})$ as $R_{s}= \frac{A_{c} - A_{p}}{A_{c}}$. Centered Kernel alignment (CKA) \cite{cka} maps illustrates model behaviours. More details in  supplementary.

\section{Benchmark Analysis}
\label{sec:benchmark}
In this section, we perform analysis across the following five aspects:

\noindent{\textit{\textbf{Effect of Pretraining Dataset Size:}}}  
In self-supervised learning, a fundamental question is whether the size of the pretraining dataset influences downstream task performance. It is crucial to determine if increasing the dataset size leads to proportional improvements in performance. Additionally, a common trend in SSL is to train models for extended durations during pretraining. We investigate whether prolonged training meaningfully contributes to performance gains. To address these questions, we analyze different training stages across multiple architectures and pretext tasks.  

\noindent{\textit{\textbf{Impact of Task Complexity:}}}  
Previous studies suggest that increasing task complexity enhances representation learning, whereas reducing complexity may lead to suboptimal solutions. We examine this hypothesis in detail by evaluating multiple tasks across different model architectures to assess how complexity influences learned representations.  

\noindent{\textit{\textbf{Effect of Data Distribution:}}}  
Most existing self-supervised methods evaluate performance on Kinetics-400 (K400) and UCF101, both of which exhibit strong appearance bias. However, we focus on datasets where temporal dynamics play a crucial role, such as Something-Something v2 (SSv2) and Diving48. This allows us to better understand the generalization of self-supervised methods beyond appearance-driven cues.  

\noindent{\textit{\textbf{Robustness of SSL Tasks:}}}  
We assess the robustness of SSL methods against data noise \cite{noise}, examining which factors contribute to their stability under domain shifts. Understanding these aspects is critical for developing more resilient self-supervised learning approaches.  

\noindent{\textit{\textbf{Feature Analysis:}}}  
Finally, we analyze the learned feature space to determine whether representations are complementary when models are trained under different protocols. This helps uncover how different pretext tasks and training conditions influence feature learning and transferability.

 \begin{figure}[t!]

      \includegraphics[width=\linewidth]{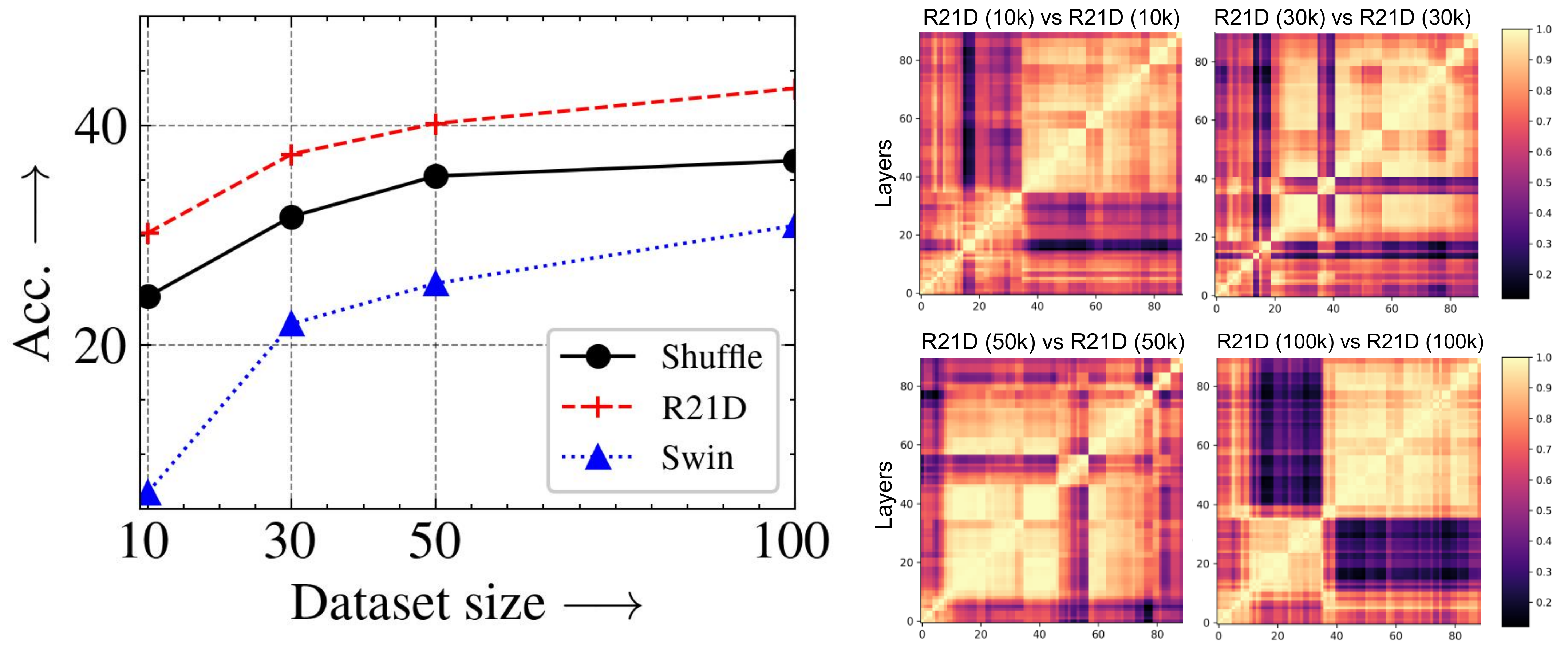}
  
    \caption{Left: \textbf{Dataset subset} performance for three different architectures on RSPNet pretext task (x-axis: subset size, y-axis: Top-1 Accuracy). Here, 10 means 10k dataset subset, 30 means 30k, and so on.  Right: \textbf{CKA maps} for RSPNet on different subsets with R21D backbone.
    }
    \label{fig:subset_analyze}
    \vspace{-10pt}
\end{figure}

\subsection{Effect of dataset-size}
We begin by analyzing the impact of pretraining dataset size variation. The network is trained on four subsets of Kinetics-400 (K400): 10k, 30k, 50k, and 100k videos, ensuring that the number of videos per class remains consistent across subsets. Each smaller pretraining dataset is a strict subset of the larger one (i.e., \(10k \subset 30k \subset 50k \subset 100k\)). We examine three key aspects related to the dependence on pretraining subset size:  
a) how different pretext tasks behave as the pretraining dataset size increases,  
b) performance variations across different backbone capacities, and  
c) the effect of training duration across different pretext tasks.  

\noindent \textbf{Observations:} As shown in Table \ref{tab:pretrain_subset_tasks}, we observe that except for TDL, all pretext tasks exhibit performance improvements with an increase in pretraining dataset size. When analyzing specific pretext task transformation categories (Table \ref{tab:pretrain_subset_tasks}), we find that \textit{spatio-temporal} tasks benefit the most from increased data, achieving a performance gain of approximately 13\%, whereas \textit{temporal} tasks show the least improvement, with only a 3\% gain. Regarding training duration (Table \ref{tab:duration_all}), performance gains become marginal (\(<\)1.5\%) after 100 epochs, indicating diminishing returns from extended training. Comparing contrastive and non-contrastive approaches, contrastive methods exhibit an average improvement of 1\% beyond 100 epochs, while non-contrastive tasks achieve a more significant gain of approximately 5%.

%  with 50k pre-training subset size 
\begin{table*}[t]
\centering
\begin{minipage}[t]{.32\textwidth}
\centering
\caption{Evaluation of different pretext tasks on \textbf{different subset size} on R21D network (\%). }
\renewcommand{\arraystretch}{1.1}
\scalebox{0.65}{
\begin{tabular}{r| ccc | ccc}
\specialrule{1.5pt}{0pt}{0pt}
\rowcolor{mygray}
\cellcolor{mygray} & \multicolumn{3}{c|}{ \cellcolor{mygray} Non-Contrastive} & \multicolumn{3}{c}{ \cellcolor{mygray} Contrastive}   \\ 
\rowcolor{mygray} 
Subset & Rot &VCOP & PRP & CVRL & TDL & RSPNet  \\
\hline \hline
10k &37.6 & 46.3  &17.5&55.9&31.1&70.9 \\
30k & 36.2 & 50.4 &42.7 &56.9  & 30.9&76.4\\
50k & 41.2 & 51.5 &46.2&61.2&30.2 &78.0\\
\specialrule{1.5pt}{0pt}{0pt}
\end{tabular}}
\label{tab:pretrain_subset_tasks}
\end{minipage}%
\hfill
\begin{minipage}[t]{.32\textwidth}
\centering
\caption{\textbf{Performance at different stages} of training for all pretext tasks on R21D (50k)(\%). }
\renewcommand{\arraystretch}{1.1}
\scalebox{0.63}{
\begin{tabular}{r | ccc | ccc}
\rowcolor{mygray} 
\specialrule{1.5pt}{0pt}{0pt}
\rowcolor{mygray}
\cellcolor{mygray} & \multicolumn{3}{c|}{ \cellcolor{mygray} Non-Contrastive} & \multicolumn{3}{c}{ \cellcolor{mygray} Contrastive}   \\ 
\rowcolor{mygray} 
Epochs & Rot & VCOP & PRP & CVRL & TDL & RSPNet  \\
\hline \hline
50   & 35.4 &52.2&24.1&55.7 &32.1& 75.0\\
100  & 37.3&52.3&34.8& 58.5&31.3& 76.1\\
150  & 40.7&51.3&46.7& 60.2&31.5 &76.5\\
200  & 40.9& 52.8 & 45.0& 60.5&30.2& 77.4\\ 
\specialrule{1.5pt}{0pt}{0pt}
\end{tabular}}
\label{tab:duration_all}
\end{minipage}%
\hfill
\begin{minipage}[t]{.32\textwidth}
\centering
\caption{\textbf{Complexity Variation.} TC: Task complexity. Results are shown on UCF101 with ShuffleNet/R21D backbone.}
\renewcommand{\arraystretch}{1.1}
\scalebox{0.7}{
\begin{tabular}{c| cccc}
\rowcolor{mygray} 
\specialrule{1.5pt}{0pt}{0pt}
TC $\downarrow$& S & T & ST \\ 
\hline \hline
C1 & 20.1/48.3 & 41.6/\textbf{56.8} & \textbf{24.2}/38.9\\
C2 & \textbf{20.2}/\textbf{58.3} & \textbf{41.8}/54.8 & 18.1/44.4\\
C3 & 16.6/41.2 & 40.6/55.6 & 21.9/\textbf{46.2}\\
% \checkmark & \checkmark & \checkmark & 41.2 & 22.1 & 31.8& 19.6 \\ 
% &&  & \improve{~+3.6} & \improve{~+2.9} & \improve{~+3.0} & \improve{~+4.3}  \\ 
\specialrule{1.5pt}{0pt}{0pt}
\end{tabular}}
\label{tab:shuffle_complexity}
\end{minipage}
\end{table*}

\noindent \textbf{Inference:} (i) \textit{Spatio-temporal pretext tasks exhibit the highest performance gains with an increase in dataset size, as they rely on transformations across both spatial (appearance) and temporal (motion) dimensions, making them more dependent on larger datasets compared to other tasks.}  
(ii) \textit{Contrastive tasks converge faster than non-contrastive ones, reaching their optimal performance in a relatively shorter training duration.}

\subsection{Impact of change in task complexity}
Next, we examine the impact of task complexity, focusing exclusively on non-contrastive tasks, as defining complexity for contrastive-based approaches is non-trivial.  We evaluate three levels of complexity (C1, C2, C3) for each task:  
a) \textit{RotNet}: Varying the number of rotation angles from 2 to 4,  
b) \textit{VCOP}: Increasing the number of shuffled clips from 3 to 5, and  
c) \textit{PRP}: Modifying the dilation sampling rates from 2 to 4 classes.  We aim to address two key questions:  
1) Does increasing task complexity lead to better spatio-temporal feature learning during pretraining?  
2) Does model capacity influence the effectiveness of task complexity?  

\noindent \textbf{Observations:} From Table \ref{tab:shuffle_complexity}, we observe that increasing task complexity does not always lead to improved performance. For example, ShuffleNet exhibits minimal improvement or even performance degradation with increased task complexity. CKA maps reveal a transition from staggered grid-like structures to multi-block patterns, indicating saturation as complexity increases. Among different transformation categories, performance gains from increased complexity are more evident in larger models, particularly for \textit{spatio-temporal} tasks. Comparing ShuffleNet and R21D, we find that R21D produces more structured feature representations, while ShuffleNet results in dark block patterns, suggesting that larger models retain the ability to learn richer representations. Additional CKA maps are provided in the supplementary materials.  

\noindent \textbf{Inference:}  
(i) \textit{An increase in pretext task complexity does not always translate to better spatio-temporal feature learning. Its effectiveness depends on both the task itself and the model’s capacity.}  
(ii) \textit{If higher complexity leads to improved feature learning, the model must also have sufficient capacity; otherwise, the task becomes too challenging, preventing the model from learning meaningful representations.}

\subsection{Effect of dataset distribution}
Shifting our focus to datasets with stronger temporal cues, we extend our experiments by pretraining on SSv2 and fine-tuning on Diving48.
We aim to address two key questions:
a) Does the categorization of pretext tasks influence performance across \textit{source} (pretraining) and \textit{target} (downstream) datasets?
b) How does the choice of \textit{source} dataset impact pretext tasks that focus exclusively on either \textit{spatial} or \textit{temporal} learning?

\begin{figure}
    \centering
\begin{subfigure}{0.42\linewidth}
\centering
\includegraphics[width=\linewidth]{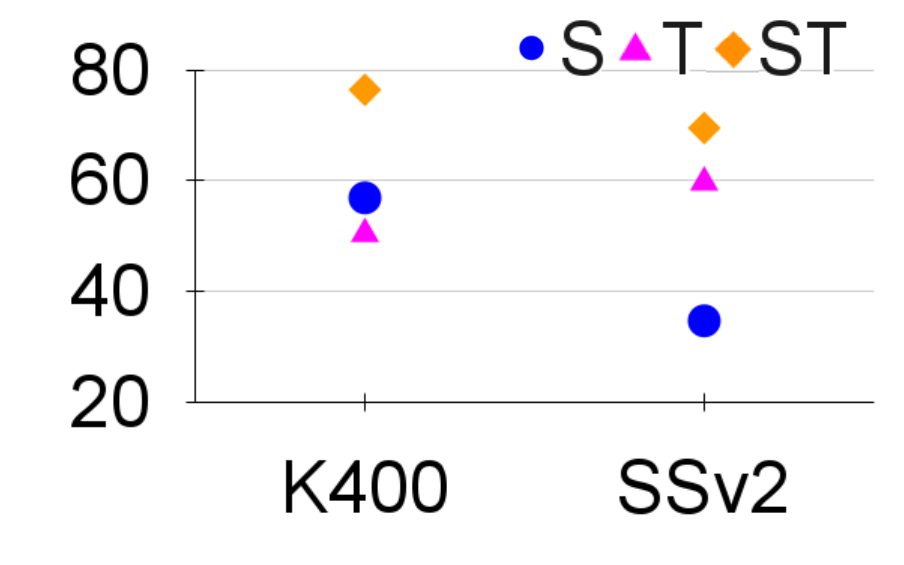}
\caption{UCF101}
\label{ood_ucf101}
\end{subfigure}
\begin{subfigure}{0.42\linewidth}
\centering
\includegraphics[width=\linewidth]{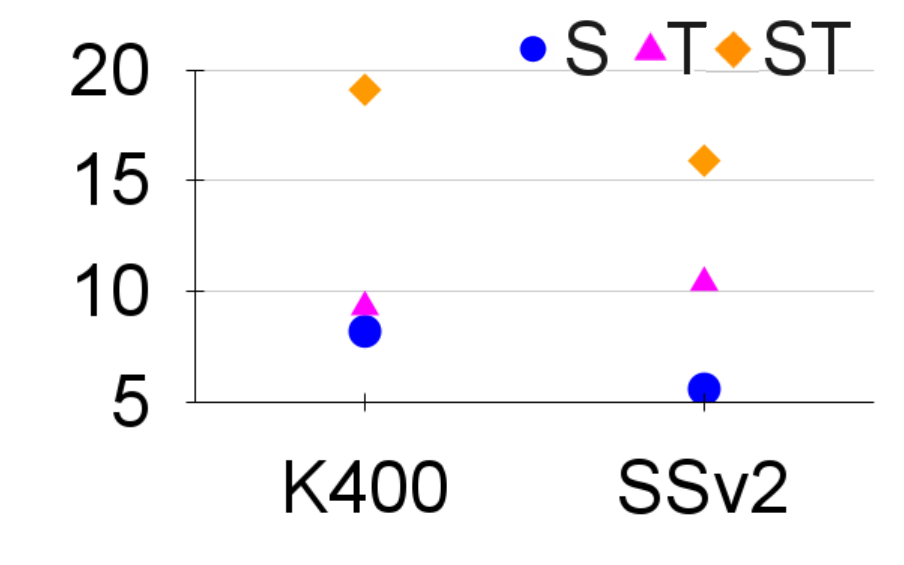}
\caption{DV48}
\label{ood_dv48}
\end{subfigure}
% % \small
\caption{\textbf{Effect of different dataset distributions:} 
% Pretraining on K400 and SSv2 with 30k subset size, finetuning on UCF101/Diving48 using R21D network. 
Here, S, T, and ST mean spatial(CVRL), temporal(VCOP), and, spatio-temporal(RSPNet) respectively. X-axis shows \textit{source} dataset and Y-axis shows Top-1 accuracy.}
\label{fig:ood_multift}
\end{figure}

\noindent \textbf{Observations:} From Figure \ref{fig:ood_multift}, we observe that \textit{spatio-temporal} pretext tasks consistently outperform others on both downstream datasets, UCF101 and Diving48, with margins of 15-40% and 10-13%, respectively, regardless of whether the \textit{source} dataset is K400 or SSv2.
Comparing spatial and temporal pretext tasks, we find that their performance is highly dependent on the \textit{source} dataset. As shown in Figure \ref{fig:ood_multift}, both tasks perform better when the \textit{source} dataset shares similar properties with the features the pretext task is designed to learn. Furthermore, spatial pretext tasks exhibit a stronger dependence on the \textit{source} dataset, as evidenced by the significant performance drop for CVRL—40\% on UCF101 and 30\% on Diving48—when pretrained on SSv2 instead of K400. In contrast, temporal pretext tasks experience smaller performance drops of 15\% and 10\%, respectively, when pretrained on K400 instead of SSv2.

\noindent \textbf{Inference:}
(i) \textit{Spatio-temporal pretext tasks learn more generalizable features, making them less dependent on source and target data distributions.}
(ii) \textit{Spatial and temporal pretext tasks are more effective when the source dataset aligns with their respective learning objectives.}
(iii) \textit{Temporal pretext tasks perform well when the target dataset has strong temporal dependencies, whereas spatial tasks rely more heavily on the source data distribution.}

\begin{table}[]
    \centering 
\renewcommand{\arraystretch}{1.06}
\scalebox{0.8}{
\begin{tabular}{r| ccc | ccc | c}
\specialrule{1.5pt}{0pt}{0pt}
\rowcolor{mygray}
\cellcolor{mygray} & \multicolumn{3}{c|}{ \cellcolor{mygray} Non-Contrastive} & \multicolumn{3}{c}{ \cellcolor{mygray} Contrastive} &   \\ 
\rowcolor{mygray} 
& Rot & VCOP & PRP & CVRL & TDL & RSP & Avg. \\
\hline
R21D & 10.7 & 19.0 & 70.1 & 78.4 & 26.7 & 68.8 & 45.6\\
Shuffle & 28.3 & 28.4 & 22.8 & 51.9 & 43.5 & 28.6 & 33.9\\
\specialrule{1.5pt}{0pt}{0pt}
\end{tabular}}
\captionof{table}{\textbf{Robustness analysis} on the relative decrease in \% performance across different pretext tasks on noisy UCF101 dataset. The performance is averaged over 4 noises. }
\label{tab:r21d_perturb_l1}
\end{table}

\subsection{Robustness of SSL tasks}
Similar to out-of-distribution (OOD) datasets, the introduction of noise alters the data distribution, impacting model performance. We evaluate models under various types of noise introduced in \cite{ood_noise} with different severity levels, using the UCF101 test dataset. Specifically, we examine four appearance-based noise types: Gaussian, Shot, Impulse, and Speckle \cite{noise}. Our analysis focuses on two key aspects:  
a) How robust are different categories of pretext tasks to noise?  
b) Does the network architecture influence robustness against noise in the dataset?  
In the main paper, we discuss results for a single severity level, while a comprehensive analysis across multiple severity levels is provided in the supplementary material.  

\noindent \textbf{Observations:} As shown in Table \ref{tab:r21d_perturb_l1}, contrastive tasks exhibit a larger relative drop in performance compared to non-contrastive tasks across both R21D and ShuffleNet backbones. Among the models analyzed, RotNet-R21D demonstrates the highest robustness, with a 10.7\% relative decrease, whereas PRP-R21D shows the highest vulnerability, experiencing a 70.1\% drop in performance.

\noindent \textbf{Inference:} \textit{Contrastive approaches are less robust to noise as compared with non-contrastive.}
 
\subsection{Feature analysis}
\begin{figure*}[t!]
     \centering
     % \small
   \includegraphics[width=\linewidth]{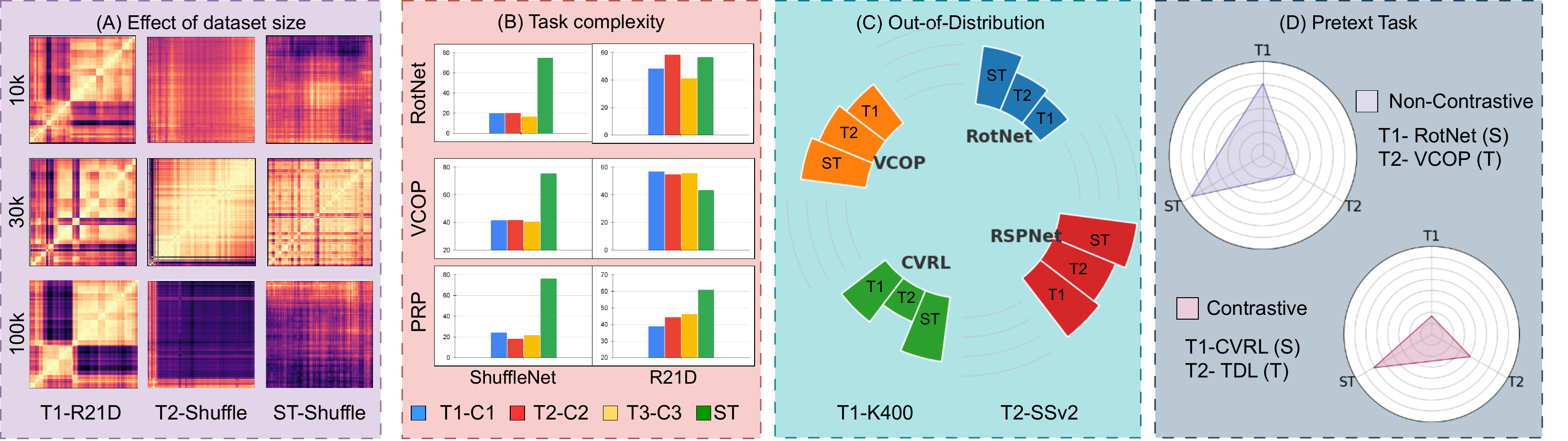}
     \caption{\textbf{Feature analysis overview.} This figure shows how KD as a tool is beneficial across multiple scenarios. Brief details for each setup (Left to right): (A) \textit{Effect of dataset size:} Teachers (T1 and T2) are different architectures for a single subset. Student model (ST-Shuffle) CKA maps shows it learns complementary information especially for 30k. (B) \textit{Task Complexity: }  Teachers are multiple complexities across the same task. (C1, C2, C3 - different complexities as teachers.) We observe in most of the scenarios, Student  (ST) networks outperforms all teacher models which proves learning of orthogonal information from multiple teachers. (C) \textit{Out-of-Distribution: } Models from different \textit{source} datasets are teachers. Student model (ST) outperforms  both teachers trained on two different datasets. (D) \textit{Pretext Tasks: } Spatial and temporal task networks are teachers, and, student model (ST) learnt from two different categories of pretext tasks - spatial and temporal incorporate knowledge from both and outperforms both of the teachers for both contrastive and non-contrastive.}
\label{fig:kd_main_fig}
\vspace{-15pt}
 \end{figure*}

We further analyze the features learned by different pretext tasks under various configurations, with a particular focus on their complementary nature. To study this, we employ knowledge distillation \cite{kdmain}, leveraging the principle that knowledge distilled from an ensemble of teacher networks enhances the performance of a student model. The multi-teacher knowledge distillation loss is defined as:  $\mathcal{L}_{KD} =  \mathcal{L}_{CE} +\mathcal{L}_{KL_{1}}+ \mathcal{L}_{KL_2}+ ... + \mathcal{L}_{KL_n} $ where \(\mathcal{L}_{CE}\) is the cross-entropy loss for hard labels, and \(\mathcal{L}_{KL_n}\) represents the KL-divergence loss from teacher \(n\). We investigate the transfer of complementary information by using our benchmark models as teachers in different combinations across four key axes:  
1) Different architectures as teachers within the same \textit{dataset size},  
2) Teachers with varying complexities within a pretext task,  
3) Models trained on multiple \textit{source} datasets, and  
4) The same architecture trained on multiple pretext tasks. Figure \ref{fig:kd_main_fig} provides a summary of the key \textit{observations} for each aspect, with further details in the supplementary material.  

\begin{figure*}[t!]
    \centering
    \begin{subfigure}{0.24\linewidth}
       \centering
      \includegraphics[width=\linewidth]{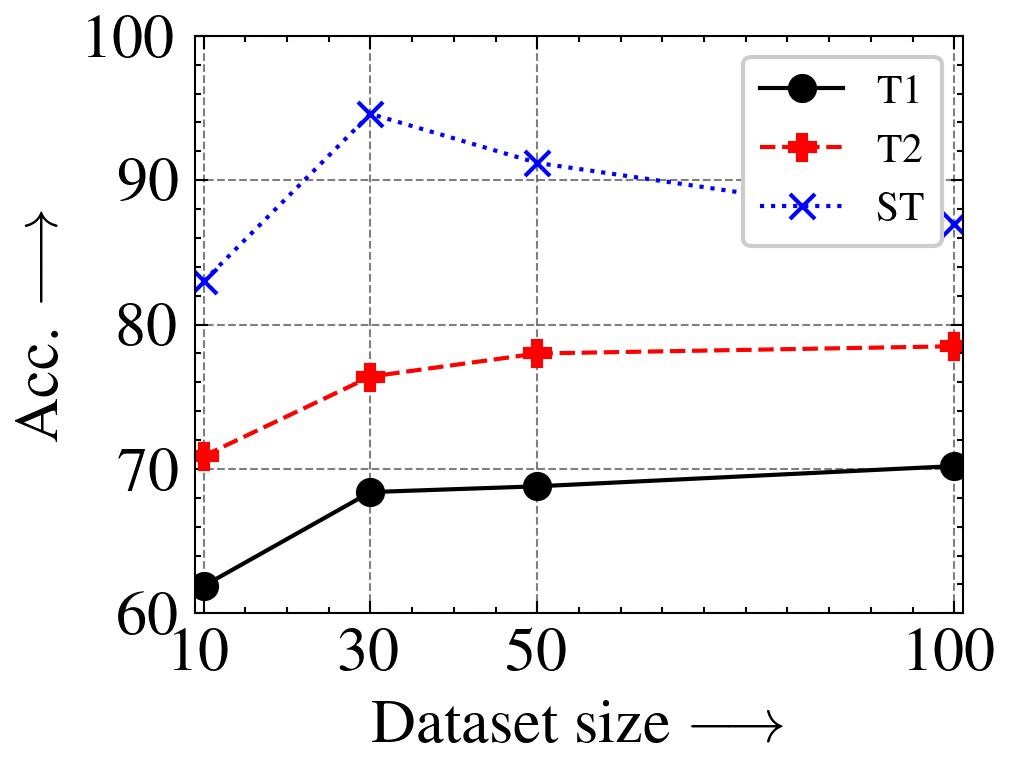}
    \caption{UCF101}
    \label{kd_ucf101}
    \end{subfigure}
    % \hfill
    \begin{subfigure}{0.24\linewidth}
       \centering
      \includegraphics[width=\linewidth]{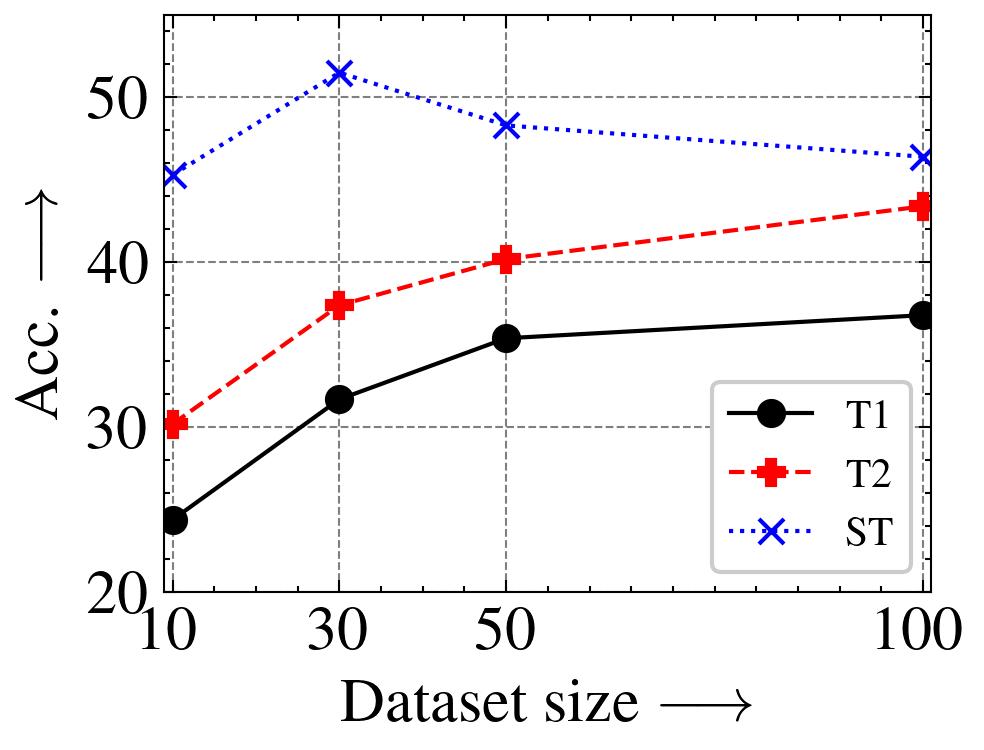}
    \caption{HMDB51}
    \label{kd_hmdb51}
    \end{subfigure}
    \centering
    \begin{subfigure}{0.24\linewidth}
       \centering
      \includegraphics[width=\linewidth]{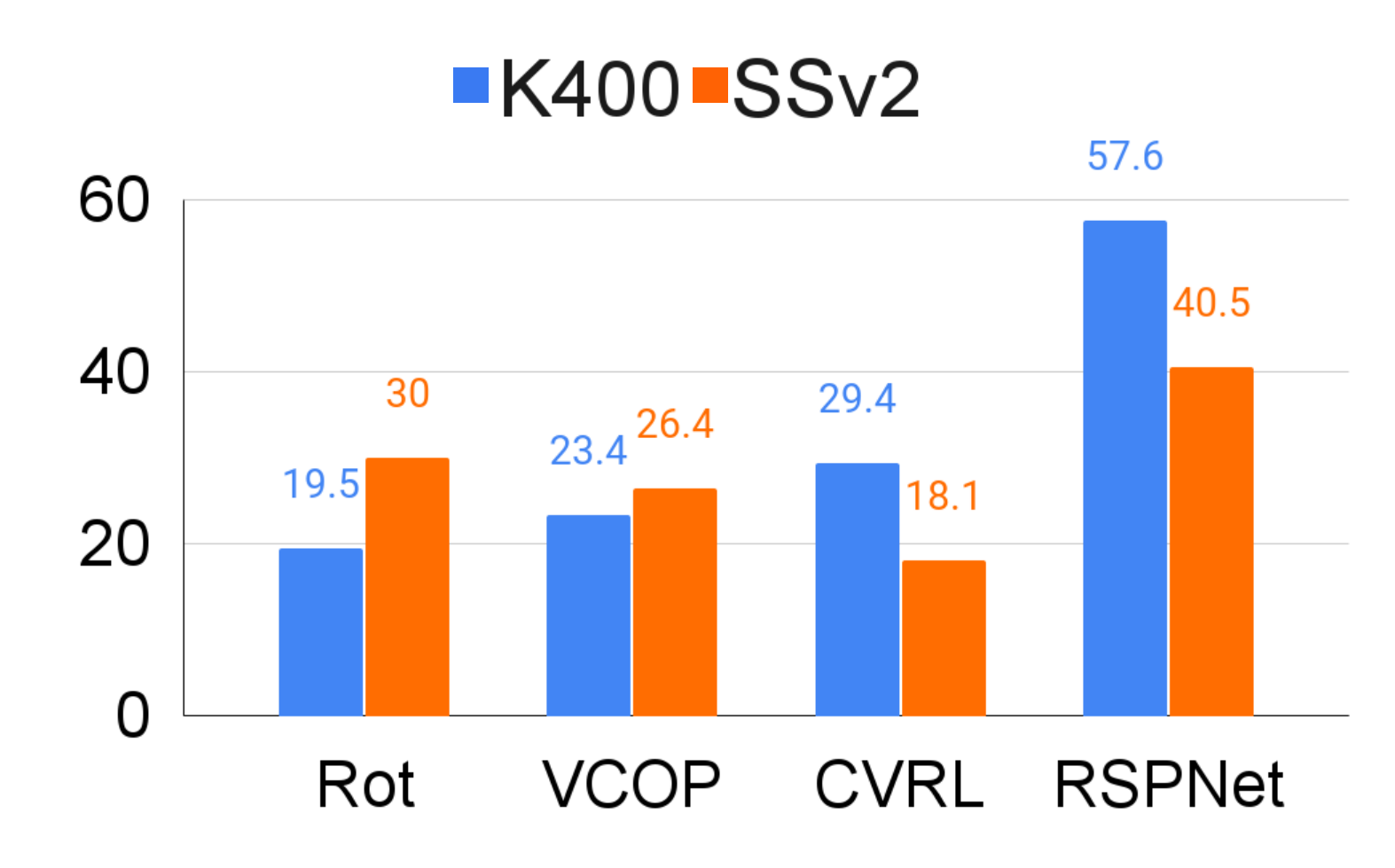}
    \caption{UCF101}
    \label{ret_ucf101}
    \end{subfigure}
    % \hfill
    \begin{subfigure}{0.24\linewidth}
       \centering
      \includegraphics[width=\linewidth]{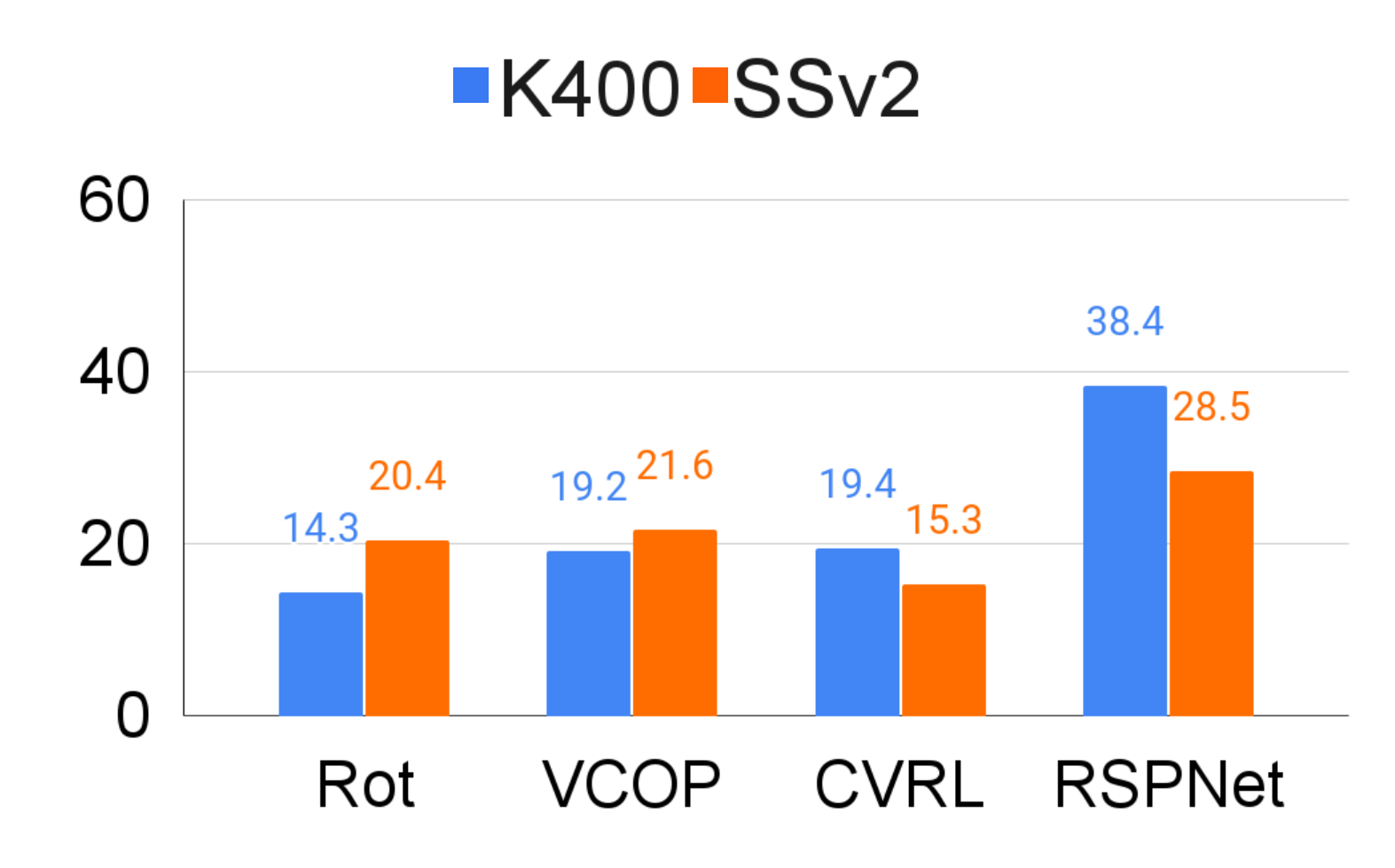}
    \caption{HMDB51}
    \label{ret_hmdb51}
    \end{subfigure}
    \caption{\textbf{Knowledge distillation} using teachers trained on multiple subset sizes on RSPNet. Student: ShuffleNet a) UCF101 and b) HMDB51. Here T1 is Teacher-1 (shufflenet) and T2 is teacher-2 (R21D). \textbf{Top@5 Clip Retrieval} - R21D on c) UCF101 and d) HMDB51, pre-trained on K400 and SSv2 - 30k subset.
    }
    \label{fig:clip_ret}
    \vspace{-10pt}
\end{figure*}

\paragraph{Observations:}  
Although teacher network performance improves with an increase in dataset subset size, the gain in complementary information diminishes beyond 30k samples (Figs. \ref{fig:clip_ret}(a) \& \ref{fig:clip_ret}(b)). However, knowledge distillation significantly reduces training time while improving performance, as seen in Fig. \ref{fig:kd_main_fig}(a). Regardless of the pretext task category, smaller architectures benefit more from complementary information and often outperform their teacher models. In contrast, larger architectures show task-dependent improvements. Moreover, across both transformation-based and contrastive tasks, knowledge distillation from multiple \textit{source} datasets enhances feature learning and results in superior student model performance. Finally, student networks consistently outperform standalone spatio-temporal models in both contrastive and non-contrastive domains.  

\paragraph{Inference:}  
(i) \textit{Knowledge can be effectively distilled across different architectures for a given dataset subset size (Fig. \ref{fig:kd_main_fig} (a))}.  
(ii) \textit{Leveraging multiple \textit{source} datasets introduces complementary information, leading to improved learning (Fig. \ref{fig:kd_main_fig} (c))}.  
(iii) \textit{Pretext tasks from different categories learn orthogonal features, contributing to a more diverse and enriched representation space (Fig. \ref{fig:kd_main_fig} (d))}.  

\section{Lessons Learned}

Our analysis across different axes reveals several key insights:  
(i) Contrastive tasks exhibit faster learning but are highly susceptible to noise.  
(ii) Increasing dataset size or task complexity does not significantly improve the learning of spatio-temporal features in smaller models; however, the features learned are more robust to noise.  
(iii) Temporal tasks are inherently more challenging to learn, as performance gains remain minimal even with increased training time, dataset size, or task complexity, indicating the intrinsic difficulty of these tasks.  
(iv) Spatio-temporal pretext tasks benefit from increased dataset size and task complexity (provided the model capacity allows), and their ability to learn meaningful spatio-temporal features remains independent of data distribution.  Building upon these findings, we further analyze the learned feature space to understand how models acquire orthogonal information within each comparison axis. Based on these insights, we investigate strategies to enhance performance on downstream tasks, specifically action classification and clip retrieval.

\begin{table*}[t!]
	\centering
        \caption{\textbf{Comparison with previous approaches} pre-trained on K400. Ours ( $^{*}$ best performing) is RSPNet pretrained on 30k subset of K400. $^{\dagger}$ - Different pre-training data. (\%) }
	\renewcommand{\arraystretch}{1.1}
	\scalebox{0.8}{
		\begin{tabular}{r| ccc | ccc }
  \specialrule{1.5pt}{0pt}{0pt}
				\rowcolor{mygray} 
        Approach&Venue & NxW/H & Backbone &  Pre-training & UCF101 & HMDB51 \\
\hline
\textbf{Generative}\\
\hline
VIMPAC \cite{Tan2021VIMPACVP}&  - & 10x256 & ViT-L & HTM & 92.7 & 65.9 \\
VideoMAE \cite{videomae} & NeurIPS'22 & 16x224& ViT-B & K400& 91.3 & 62.6\\
MME \cite{mme} & CVPR'23 & 16x224 & ViT-B & K400 & 96.5 & 78.0\\
MVD \cite{mvd} & CVPR'23 & 16x224 & ViT-B & IN1K+K400 & 97.0 & 76.4\\
EVEREST \cite{everest} & - & 16x224 & ViT-B & - & 93.4 & 68.1\\
SCE \cite{sce} & WACV'23 & 16x224 & ResNet3D-50 & K400 & 95.3 & 74.7\\
\hline
\textbf{Context} \\
\hline
% V
PacePred \cite{pace} &  ECCV'20 &  16x112& R21D-18 & K400& 77.1 & 36.6\\
TempTrans \cite{temporals} & ECCV'20 & 16x112 & R3D-18& K400 & 79.3& 49.8 \\
STS \cite{stats} &TPAMI-21 & 16x112& R21D-18 & K400& 77.8 & 40.5\\
VideoMoCo \cite{videomoco} & CVPR'21& 16x112& R21D-18 & K400& 78.7 & 49.2\\
RSPNet \cite{rspnet} & AAAI'21& 16x112& R21D-18 & K400& 81.1 & 44.6\\
TaCo \cite{taco} & - & 16x224& R21D-18 & K400& 81.8 & 46.0\\
TCLR\cite{tclr} & CVIU'22& 16x112 & R21D-18 & K400& 88.2 & 60.0\\
% CVRL$^{\dagger}$ \cite{cvrl} & CVPR'21& 32x224& R21D-18 & K400& 92.9 &67.9\\
CVRL \cite{cvrl} & CVPR'21& 32x224& R21D-18 & K400& 92.9 &67.9\\
TransRank \cite{transrank}&  CVPR'22 & 16x112 & R21D-18& K200& 87.8 & 60.1 \\
\hline
\textbf{Multi-Modal}\\
\hline
AVTS \cite{avts}& NeurIPS'18 &  25x224 & I3D & K400& 83.7 & 53.0 \\
GDT \cite{gdt} $\dagger$ & - & 32x112& R21D & IG65M & 95.2 & 72.8\\
XDC \cite{xdc} & NeurIPS'20& 32x224& R21D &K400 & 84.2 & 47.1\\
\hline
Ours $^{*}$ & - &16x112& R21D-18 &K400-30k & 97.3 & 51.5\\
                \specialrule{1.5pt}{0pt}{0pt}
		\end{tabular}}
	\label{tab:action_recog_acc}
 \vspace{-15pt}
\end{table*}

\noindent{\textbf{Clip retrieval}} For this downstream task, we generate feature vectors using pretrained weights and retrieve the nearest neighbor by computing the cosine distance between test and train feature vectors. We conduct our analysis on UCF101 and HMDB51 with different \textit{source} data distributions: K400 and SSv2. \textbf{\textit{Observations:}} The spatio-temporal pretext task consistently outperforms other categories, independent of the \textit{source} data distribution, aligning with our earlier findings. Contrastive learning excels at capturing \textit{appearance} features during pretraining, which is advantageous since both downstream datasets are \textit{appearance}-centric. Temporal tasks exhibit comparable performance regardless of the \textit{source} dataset used for pretraining. This suggests that, even when trained on an appearance-based dataset, temporal tasks do not overly rely on spatial features.

\noindent{\textbf{Action Classification}} For this task, we fine-tune the model end-to-end on the UCF101 and HMDB51 datasets. As shown in Table \ref{tab:action_recog_acc}, our best-performing model is obtained through knowledge distillation, as discussed in the previous section. The results demonstrate that our model surpasses prior state-of-the-art methods. \textbf{\textit{Observations:}} Despite being trained on only 30k videos—significantly fewer than the 200k+ videos used by other pretext tasks—our model achieves superior performance on UCF101, outperforming both single- and multi-modal approaches by a substantial margin. On HMDB51, our model achieves competitive performance with a classification accuracy of 51.5\%.

\subsection{Surprising Findings}
We have multiple inferences from different axes of analysis. However, to club a few which are new and helpful for video self-supervised community, we list down those here:

\begin{table*}[t]
	\centering
	\begin{minipage}{.48\textwidth}
		\centering
             \caption{\textbf{Analysis on ViFMs}.  Zero-shot classification accuracy on UCF-101. I:Image, V: Video.}
		\renewcommand{\arraystretch}{1.1}
		\scalebox{0.68}{
			\begin{tabular}{r | ccc | ccc}
   \rowcolor{mygray} 
				\specialrule{1.5pt}{0pt}{0pt}
                    \rowcolor{mygray} 
 ViFM & Type. &  Pretraining Data & Frames x Rate& Accuracy  \\ 
 \hline \hline
 %VideoMAE \cite{videomae} & video-based & Generative & Kinetics-400 & 16 x 4 & 76.940\\ 
 ViFi-CLIP \cite{vificlip} & I  & K-400 & 32 x 2 & 77.3\\ 
 X-CLIP \cite{xclip} & I  & K-400 & 8 x 8 & 71.6\\ 
 EZ-CLIP \cite{ezclip} & I  & K-400 & 8 x 8 & 70.5\\ 
 ViCLIP \cite{viclip}& V  & InternVid-10M & 8 x 8 & 75.5\\ 
 LanguageBind \cite{languagebind} & V  & VIDAL-10M & 8 x 8 & 69.9\\  \specialrule{1.5pt}{0pt}{0pt}
		\end{tabular}}
		\label{tab:vifm}
	\end{minipage}%
	\hfill
        \begin{minipage}{.5\textwidth}
		\centering
            \caption{\textbf{Knowledge Distillation} between different ViFM pairs as teachers, and R21D as the student.}
		\renewcommand{\arraystretch}{1.1}
		\scalebox{0.68}{
			\begin{tabular}{c| ccccc}
				\rowcolor{mygray} 
				\specialrule{1.5pt}{0pt}{0pt}
                 ViFM & X-CLIP & ViFi-CLIP & EZ-CLIP & ViCLIP & LanguageBind\\ 
 \hline \hline
 %VideoMAE & X & 81.822 & 81.499 & 86.725 & 87.296 & 87.953\\ 
 X-CLIP & X & 83.2 & 88.7 & 88.2 & 87.6\\
 ViFi-CLIP & X & X & 88.0 & 86.6 & 86.6\\
 EZ-CLIP & X & X & X & 85.0 & 86.9\\
 ViCLIP & X & X & X & X & 85.4\\ 
 LanguageBind & X & X & X & X & X\\
          \specialrule{1.5pt}{0pt}{0pt}
		\end{tabular}}
		\label{tab:vifmkd}
	\end{minipage}
 \vspace{-10pt}
\end{table*}

\noindent{\textit{\textbf{Dataset Size and Training Time Dependency:}}} Contrary to the conventional belief that extensive training data is essential for achieving optimal performance, our findings reveal that beyond a certain threshold, additional data yields diminishing returns for self-supervised learning (SSL). This insight has significant implications, as it enables a substantial reduction in training data requirements while achieving nearly a 10× reduction in training time—particularly beneficial for computationally intensive video processing tasks. Moreover, we demonstrate that knowledge distillation (KD) can outperform the original approach trained on 100\% of the data while utilizing only 10\% of it, thereby optimizing resource utilization by approximately 80\%.  

\noindent{\textit{\textbf{Robustness to Real-World Noise:}}} Surprisingly, contrastive tasks exhibit greater susceptibility to noise compared to non-contrastive tasks. Additionally, in certain scenarios, smaller networks demonstrate higher robustness than larger networks. We believe these findings are novel, as no prior work in the community has systematically explored these aspects. This insight is particularly valuable for real-world deployments where robustness is a critical requirement.  

\noindent{\textit{\textbf{Complementary Knowledge:}}} The performance improvements achieved through KD from different data distributions and pretext task categories highlight a promising direction for a new SSL paradigm. Specifically, a multi-teacher, multi-student framework, where each teacher specializes in either spatial or temporal tasks and is trained on a diverse set of data sources, could provide an effective learning strategy. Our analysis suggests that such an approach would enhance feature learning and task generalization.  

\subsection{Recommendations}  
Based on our findings, we propose the following recommendations for designing effective SSL strategies:  
\begin{enumerate}  
    \item \textit{Training Speed:} If pretraining time is a constraint, contrastive tasks can accelerate the learning process. However, they may be less robust to noise.  
    \item \textit{Data Distribution:} When possible, a spatio-temporal pretext task should be preferred, as it generalizes well across data distributions. If this is not feasible, the pretext task should align with the characteristics of the pretraining dataset.  
    \item \textit{Model Capacity:} If the model has limited capacity, increasing the pretraining dataset size or using complex pretext tasks will not yield significant benefits.  
    \item \textit{Robustness:} For applications requiring high robustness, non-contrastive pretext tasks should be favored over contrastive ones.  
    \item \textit{Performance:} Pretext tasks learn complementary features across model architectures, pretraining datasets, task categories, and task complexities. This complementary knowledge can be distilled to develop stronger spatio-temporal feature representations.  
\end{enumerate}  

\subsection{Extension of Findings to Video Foundation Models (ViFMs)}  
We extend our study to Video Foundation Models (ViFMs) (Tables \ref{tab:vifm} and \ref{tab:vifmkd}). Our analysis includes both image-based ViFMs—such as EZ-CLIP, X-CLIP, and ViFi-CLIP—which are adapted from image foundation models, as well as video-based ViFMs—such as LanguageBind and ViCLIP—which are trained from scratch on videos. Notably, all ViFMs utilize contrastive pretraining objectives.

\noindent{\textit{\textbf{Dataset Size:}}} Increasing dataset size or complexity does not necessarily enhance the ability of smaller models to learn better spatio-temporal features (Table \ref{tab:vifm}). Notably, ViCLIP and LanguageBind, despite being pretrained on significantly larger datasets, underperform compared to models trained on the smaller Kinetics-400 dataset. Furthermore, simply increasing the number of frames during training yields better performance than relying solely on larger pretraining datasets.  

\noindent{\textit{\textbf{Complementary Knowledge:}}} The performance improvements observed with KD from different ViFMs suggest a promising strategy for training a new foundational model. This approach involves a multi-teacher, multi-student framework, where each teacher is a ViFM pretrained under different conditions—varying data sources, multi-stage pretraining strategies, and distinct pretraining objectives. Our analysis (Table \ref{tab:vifmkd}) indicates that such a framework would foster a more effective learning process.  

\vspace{-8pt}
\section{Conclusion}  
\label{sec:conclusion}  
In this study, we systematically analyze various factors influencing self-supervised learning in the video domain. We establish a benchmark that provides an intuitive categorization of pretext tasks, enabling a more structured comparison across different learning paradigms. To the best of our knowledge, such an in-depth exploration of self-supervised learning for video understanding has not been previously conducted. Our findings uncover several key insights that pave the way for future research directions. Additionally, we demonstrate the practical impact of these insights by achieving state-of-the-art performance on video action recognition while utilizing only 10\% of the pretraining dataset compared to existing methods. We believe this benchmark study will serve as a valuable resource for the research community, fostering a deeper understanding of self-supervised learning in the video domain.  
% \fi

{
    % \small
    \bibliographystyle{ieeenat_fullname}
    \bibliography{main}
}

\clearpage
\setcounter{page}{1}
\renewcommand{\thesection}{\Alph{section}}
\setcounter{section}{0}

\maketitlesupplementary

% \setcounter{section}{0}
% \setcounter{figure}{0}
% \setcounter{table}{0}

% \input{X_suppl}

Here, we explain things in details about pretext task, architecture setup, provide some more results and include more visual analysis. We also include tables which we were not able to include in main paper due to space limitations.

\begin{itemize}

\item Section~\ref{sec:challenge}: describes challenges and future work based on our study.

\item Section~\ref{sec:pretext}: Pretext tasks explanation used in our analysis.
\item Section~\ref{sec:implement}: Training details about architectures, datasets, and, other hyperparameters. 

\item Section~\ref{sec:add_results}: We show additional CKA maps, more results on HMDB51 dataset and more analysis on noise robustness. We added some tables for Knowledge distillation experiments that were promised in the main paper.

\item Section~\ref{sec:maintable}: We extend the main table and compare with previous state-of-the-art results on HMDB51 dataset. 
\end{itemize}

\section{Challenges and future work}
\label{sec:challenge}

There are several key challenges in video SSL and we believe 1) long-term video understanding, 2) multi-modal learning, and 3) robust learning are some of the less studied aspects. 
The novel insights in our study regarding training dataset size, model architectures, and robustness will play a crucial role in guiding future work on these research directions. 

\section{Pretext Tasks Details}
\label{sec:pretext}

In this section, we go through each pretext task in more detail that are used in our main work for analysis.

\subsection{Spatial Transformation}
\noindent \paragraph{Rotation Net \cite{rotnet} (RotNet)}  applies geometrical transformation on the clips. The videos are rotated by various angles and the network predicts the class which it belongs to. Since the clips are rotated, it helps the network to not converge to a trivial solution. 
\noindent \paragraph{Contrastive Video Representation Learning \cite{cvrl} (CVRL)} technique applies temporally coherent strong spatial augmentations to the input video. The contrastive framework brings closer the clips from same video and repels the clip from another video. With no labels attached, the network learns to cluster the videos of same class but with different visual content.

\subsection{Temporal Transformation}

\noindent \paragraph{Video Clip Order Prediction \cite{vcop} (VCOP)}  learns the representation by predicting the permutation order. The network is fed N clips from a video and then it  predicts the order from N! possible permutations. 

\noindent \paragraph{Temporal Discriminative Learning \cite{tdl} (TDL)}   In contrast to CVRL, TDL works on temporal triplets. It looks into the temporal dimension of a video and targets them as unique instances. The anchor and positive belongs to same temporal interval and has a high degree of resemblance in visual content compared to the negative.

\subsection{Spatio-Temporal Transformation}

\noindent \paragraph{Playback Rate Prediction \cite{prp} (PRP)} has two branch, generative and discriminative. Discriminative focuses on the classifying the clip's sampling rate, whereas, generative reconstructs the missing frame due to dilated sampling. Thus, the first one concentrates on temporal aspect and second one on spatial aspect. 

\noindent \paragraph{Relative Speed Perception Network \cite{rspnet} (RSPNet) } applies contrastive loss in both spatial and temporal domain. Clips are samples from a same video to analyze the relative speed between them. A triplet loss pulls the clips with same speed together and pushes clips with different speed apart in the embedding space. To learn spatial features, InfoNCE loss \cite{infonce} is applied. Clip from same video are positives whereas clips from different videos are negatives. 

\noindent \paragraph{Video MAE \cite{videomae} (V-MAE)} applies a spatio-temporal tube masking to the input video. The pretext task is to reconstruct those missing tubes. Mean-squared error loss is applied between the masked tokens and the reconstructed tokens.

\section{Implementation Details}
\label{sec:implement}

\subsection{Architecture Details}

Preliminary research has shown that 3D networks \cite{r21d, resnet3d} have outperformed 2D CNN variants on video recognition tasks. We looked into three types of capacity - small, medium and big on the basis of number of trainable parameters. The architecture details of all networks are mentioned in supplementary.

\noindent \textbf{Small capacity networks: } are resource efficient, implying they can be trained in larger batches within short span of time. The network selection is done on the basis of supervised training scores on Kinetics\cite{kinetics} and UCF101\cite{efficient3d}. ShuffleNet V1 2.0X \cite{shufflev1} utilizes point-wise group convolutions and channel shuffling. SqueezeNet \cite{squeezenet} reduces the filter size and input channels to reduce the number of parameters. MobileNet \cite{mobnetv2} has ResNet like architecture. With its depthwise convolution, there's a reduction in model size and the network can go more deep. 

% Amongst these three, we have shown results for ShuffleNet across all architectures, since, it has the best performance. 

\noindent \textbf{Medium capacity networks: } Following the conventional 3D architectures for self-supervised learning approaches C3D, R21D and R3D are used in this study. 

% We show majority our results on R21D. \YSR{why? do we need to discuss this here?}

\begin{figure*}[t]
    \centering
    \includegraphics[width=\linewidth]{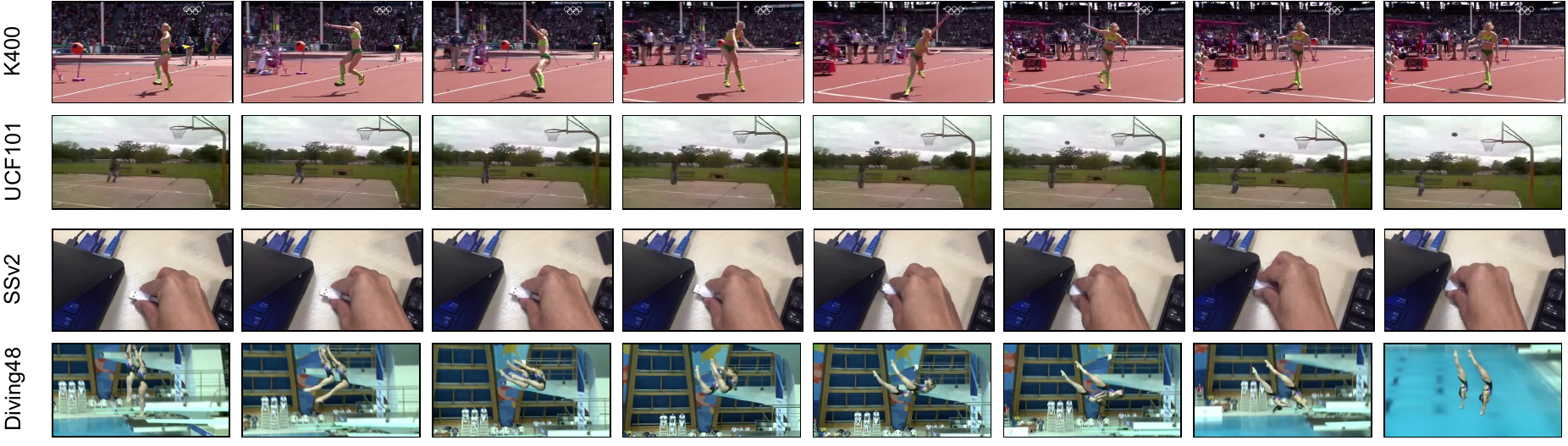}
    \caption{\textbf{Example} sample from each dataset.}
    \label{fig:viz_data}
\end{figure*}

\begin{figure*}[t!]
    \centering
    \includegraphics[width=0.9\linewidth]{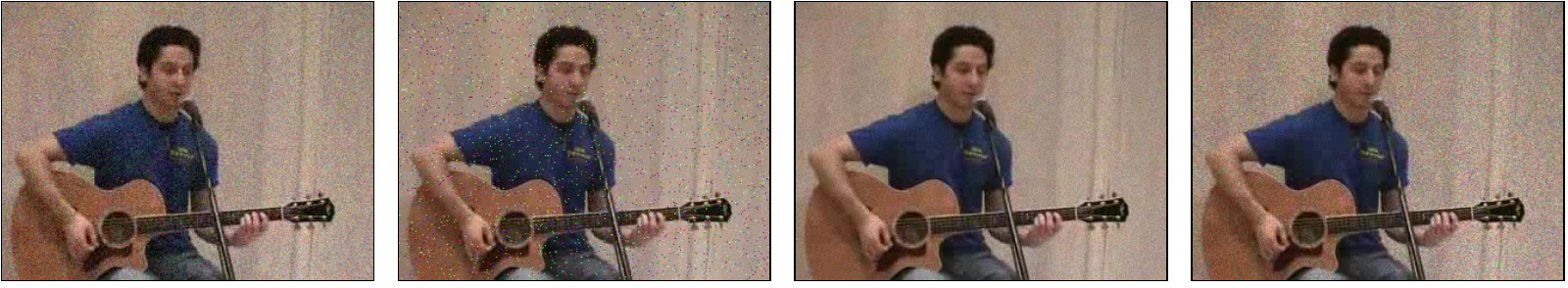}
    \caption{\textbf{Example frame sample for each noise} Gaussian, Impulse, Shot and Speckle from left to right. Sample clips are provided in supplementary. }
    \label{fig:viz_noise}
\end{figure*}

\noindent \textbf{Big Capacity networks: } Comparing across four transformer architectures, ViViT \cite{vivit} Timesformer \cite{timesformer}, VideoSwin \cite{videoswin} and MViT \cite{mvit}, we selected VideoSwin, because it outperforms others on Kinetics 400 dataset.

% firstly, it's a direct extension of ViT \cite{vit} from images to videos incorporating spatio-temporal attention, and, secondly, all these architectures have comparable performance. 

Based on \cite{efficient3d}, we probed into the performance comparison of several versions of these architectures. We choose 3D-ShuffleNet V1 2.0X, 3D-SqueezeNet, and 3D-MobileNet V2 1.0X networks based on their performance on Kinetics and UCF-101 dataset

\noindent \textbf{3D-ShuffleNet V1 2.0X \cite{shufflev1}:} It utilize point-wise group convolutions and channel shuffling and has 3 different stages. Within a stage, the number of output channel remains same. As we proceed to successive stage, the spatiotemporal dimension is reduced by a factor of 2 and the number of channels are increased by a factor of 2. V1 denotes version 1 of ShuffleNet and 2.0X denotes the 2 times number of channels compared to original configuration.

\noindent \textbf{3D-SqueezeNet \cite{squeezenet}: } It uses different alteration to reduce the number of parameters as compared to the 2D version which employs depthwise convolution. Those three modifications are: 1) Change the shape of filters from 3x3 to 1x1, 2) Input channels to 3x3 filters is reduced, and, 3) to maintain large activation maps high resolution is maintained till deep layers.

\noindent \textbf{3D-MobileNet V2 1.0X \cite{mobnetv2}: } This network employs skip connections like ResNet architecture in contrast to version 1. It helps the model in faster training and to build deeper networks. There are also linear bottlenecks present in the middle of layers. It helps in two ways as we reduce the number of input channels: 1) With depthwise convolution, the model size is reduced, and 2) at inference time, memory usage is low. V2 denotes version 2 of mobilenet and 1.0X uses the original parameter settings.

The architectures of medium capacity networks are described as follows:

\noindent \textbf{C3D \cite{c3d}: } This follows a simple architecture where two dimensional kernels have been extended to three dimensions. This was outlined to capture spatiotemporal features from videos. It has 8 convolutional layers, 5 pooling layers and 2 fully connected layers.

\noindent \textbf{R3D \cite{resnet3d}: } The 2D CNN version of ResNet architecture is recasted into 3D CNNs. It has skip connections that helps make the gradient flow better as we build more deeper networks.

\noindent \textbf{R(2+1)D \cite{r21d}: } In this architecture, 3D convolution is broken down into 2D and 1D convolution. 2D convolution is in spatial dimension and 1D convolution is along the temporal dimension. There are two benefits of this decomposition: 1) Increase in non-linearity as the number of layers have increased, and, 2) Due to factorization of 3D kernels, the optimization becomes easier. 

\noindent \textbf{VideoSwin \cite{videoswin}} It is an inflated version of original Swin \cite{Liu2021SwinTH} transformer architecture. The attention is now spatio-temporal compared to previous which is only spatial. 3D tokens are constructed from the input using patch partition and sent to the network. The architecture includes four stages of transformer block and patch merging layers.

\subsection{Original and Noise Datasets}

We have shown the examples of each dataset used in the paper in Fig. \ref{fig:viz_data}. 

The test datasets have different number of videos for different levels and types of noises. For Gaussian noise, we manipulated all 3783 samples. For noise level 1, apart from Gaussian, we had roughly 400 samples and all other levels of severity, we have approximately 550 samples. An example of each type of noise is shown in Fig. \ref{fig:viz_noise}.

\subsection{Pretext Tasks Configurations}

Here, we briefly describe the configurations used in our training. For VCOP, RotNet and PRP, we just manipulated the type of augmentation from the original work.  We applied Random Rotation, Resizing, Random Crop, Color Jittering and Random Horizontal Flipping to the input clip.  CVRL has some extra data augmentation compare to the previous ones we mentioned. It includes grayscale and gamma adjustment as well. RSPNet also uses some temporal augmentation.  For finetuning the augmentations are Resize and Center Crop for all the approaches.

The k-value for Momentum contrastive network is 16384 for RSPNet, it's 500 for TDL.

\subsection{Datasets}

Here we discuss datasets in detail. We use Kinetics-400 (K400) \cite{kinetics}  and Something-Something V2 \cite{ssv2} for our pre-training. For the downstream task evaluation, we perform our experiments on UCF-101 \cite{ucf101}, HMDB-51 \cite{hmdb51}, and Diving48 \cite{dv48}. Since, the pretraining and finetuning datasets are different, the performance variation will provide us a better picture about how much meaningful spatiotemporal features are learned by these networks. K400 has approximately 240k videos distributed evenly across 400 classes respectively. The approximate number of videos in finetuning datasets are: 1) UCF101-10k, 2) HMDB51-7k, and, 3)Diving48-16k. The datasets can be categorized into two ways:

\noindent \textbf{Appearance-based: } Kinetics, UCF101 and HMDB51 comes under this category \cite{app1, app2}.  Kinetics videos length are generally 10s centered on human actions. It mainly constitutes singular person action, person-to-person actions and person-object action. For pre-training, we select a random subset of videos and maintain equal distribution from each class. Unless otherwise stated, pre-training is done on K400-50k subset  for all experiments.

\noindent \textbf{Temporal-based: } In Kinetics, we can estimate the action by looking at a single frame \cite{app1, app2}. From Fig. \ref{fig:viz_data}, top two rows, we can see the person with a javelin and basketball. This information helps in class prediction. Looking at bottom two rows (SSv2 and Diving48 respectively), we can't describe the activity class until we look into few continuous frames. It shows that temporal aspect plays an important role for these datasets, that's why we categorize them into temporal-based datasets. 

\noindent \textbf{UCF-101 \cite{ucf101} :} It's an action recognition dataset that spans over 101 classes. There are around 13,300 videos, with 100+ videos per class. The length of videos in this dataset varies from 4 to 10 seconds. It covers five types of categories: human-object interaction, human-human interaction, playing musical instruments, body motion and sports.

\noindent \textbf{HMDB-51 \cite{hmdb51} :} The number of videos in this dataset is 7000 comprising 51 classes. For each action, at least 70 videos are for training and 30 videos are for testing. The actions are clubbed into five categories: 1) General facial actions, 2) Facial actions with body movements, 3) General body movements, 4) Body movements with object interaction, and, 5) Body movements for human interaction. 

\section{Additional Results}
\label{sec:add_results}

Here, we will talk about some additional results, to further strengthen the claims made in the main paper. 

\subsection{Preliminary Experiments}

\begin{figure*}[t!]
     \centering
   \includegraphics[width =0.8\linewidth]{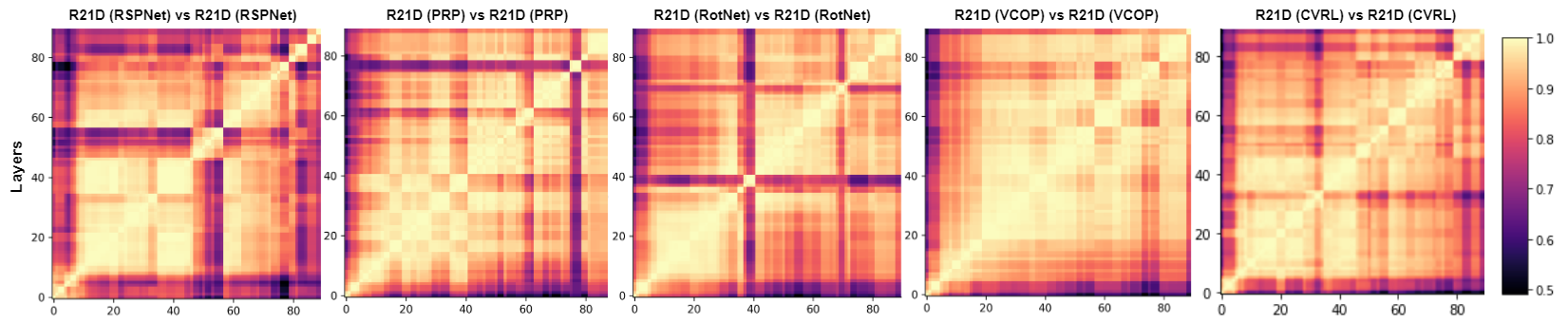}
     \caption{\textbf{Pretext tasks CKA maps} for RSPNet, PRP, RotNet, VCOP, CVRL on K-400 50k  subset using R21D network (Left to right). R21D pretrained on K400 shows a semi-block structure for VCOP, indicating near-saturation condition of the network on this pretext task. It shows a more prominent grid-based structure on CVRL and RSPNet instead. These observations corroborate the quantitative results, where pretraining on K400 for both CVRL and RSPNet gives better performance.
     }
\label{fig:cka_r21dpretext}
 \end{figure*}
  \begin{figure*}[t!]
     \centering
   \includegraphics[width =0.8\linewidth]{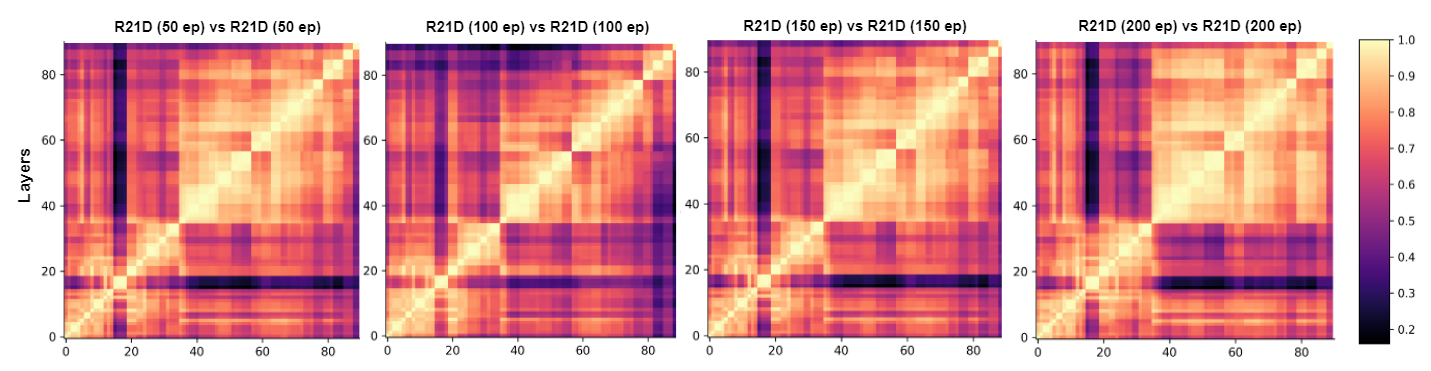}
     \caption{\textbf{Training time CKA maps} on 50, 100, 150, 200 epochs of R21D network on RSPNet pretext for K-400 10k subset (Left to right). The block structure is visible from 50 epochs itself, which then darkens and becomes prominent by 200 epochs. With 10k subset, the saturation starts hitting at 100 epochs.
     }
\label{fig:cka_10k_epoch}
 \end{figure*}
     
\paragraph{Pretext tasks evaluation} Figure \ref{fig:cka_r21dpretext} depicts the hidden representations of R21D network pretrained on different pretext tasks. Here the 50k subset of K-400 was used for pretraining, and finetuned on UCF-101. 

\paragraph{Linear Probing vs Finetuning} Firstly, we discuss linear probing (LP) vs finetuning (FT) results for different pretext tasks and different architectures. From Table \ref{fp_lt}, we can see that FT outperforms LP by a margin of approximately 20\% and 40\% on ShuffleNet and R21D respectively. Thus, we perform finetuning for all of our analysis.

\paragraph{Network Parameters} We have shown the performance across different architectures in Table \ref{tab:gflops}. ShuffleNet and R21D performs the best across small and medium capacity networks in most of the pretext tasks. Thus, we choose ShuffleNet and R21D for our benchmark analysis.

\begin{table}[t!]
        \centering
        \begin{tabular}{c ccc ccc}
        \toprule
        & \multicolumn{3}{c}{Non-contrastive} & \multicolumn{3}{c}{Contrastive} \\
        \cmidrule(lr){2-4} \cmidrule(lr){5-7}
        Epochs & VCOP & Rot& PRP & CVRL & TDL & RSPNet\\
        \midrule
        10k  & 18.9&15.0&9.2&22.2 &9.9 &30.2\\
        30k  & 19.3&11.7&11.5& 25.0 & 10.1& 37.3\\
        50k  & 17.3&12.2&10.2&29.3&9.5&40.2\\
        \bottomrule
        \end{tabular}
        \caption{\textbf{Evaluation of different pretext tasks} on different subset size on R21D network on HMDB51 dataset.}
     \label{tab:pretrain_subset_tasks_full}
     
\end{table}

\begin{table}[t!]
\centering
\begin{tabular}{c cc ccc}
\toprule
 Network & LP & FT & RotNet & VCOP & PRP\\
\midrule
\multirow{2}{*}{Shuffle} & \checkmark & & 4.3 & 12.3 & 2.8 \\
& & \checkmark & 16.6 & 40.8& 21.9 \\
\midrule
\multirow{2}{*}{R21D} & \checkmark & & 2.7 & 12.2 & 4.6\\
& & \checkmark & 41.2 & 51.5 & 46.2\\
\bottomrule
\end{tabular}
\caption{\textbf{Downstream accuracy} classification on UCF-101 dataset. FT: Finetuning LP: Linear Probing}
\label{fp_lt}
\end{table}

\begin{table*}[t!]
\centering
\begin{tabular}{c c c c ccc}
\toprule
 Networks & Parameters & GFLOPs  & Rot$^{\dagger}$ & VCOP $^{\dagger}$ & PRP$^{\dagger}$ & RSPNet\\
\hline
ShuffleNet & 4.6M & 1.1 & \textbf{42.2} & \textbf{41.6} & \textbf{41.1} & \textbf{68.8}\\
MobileNet & 3.1M  & 1.1 & 38.0 & 40.0 & 37.4 &63.1\\
SqueezeNet & 1.9M & 1.8 & 41.3 & 41.4 &39.2 & 62.9\\
\midrule
C3D & 27.7M  & 77.2& \textbf{57.7} & 54.5& 58.1 &67.6\\
R3D & 14.4M &  39.8 & 51.1 & 50.7 &52.1 &62.1\\
R(2+1)D & 14.4M & 42.9 & 46.9 & \textbf{56.8}& \textbf{58.9} &\textbf{78.0}\\
\bottomrule
\end{tabular}
\caption{\textbf{Comparison of FLOPs} and trainable parameters for each network on UCF101 dataset. $^{\dagger}$ - pretraining on Kinetics 700 \cite{k700}.}
\label{tab:gflops}
\end{table*}

\subsection{Effect of dataset size}
\begin{figure}[t!]
    \centering
      \includegraphics[width=0.8\linewidth]{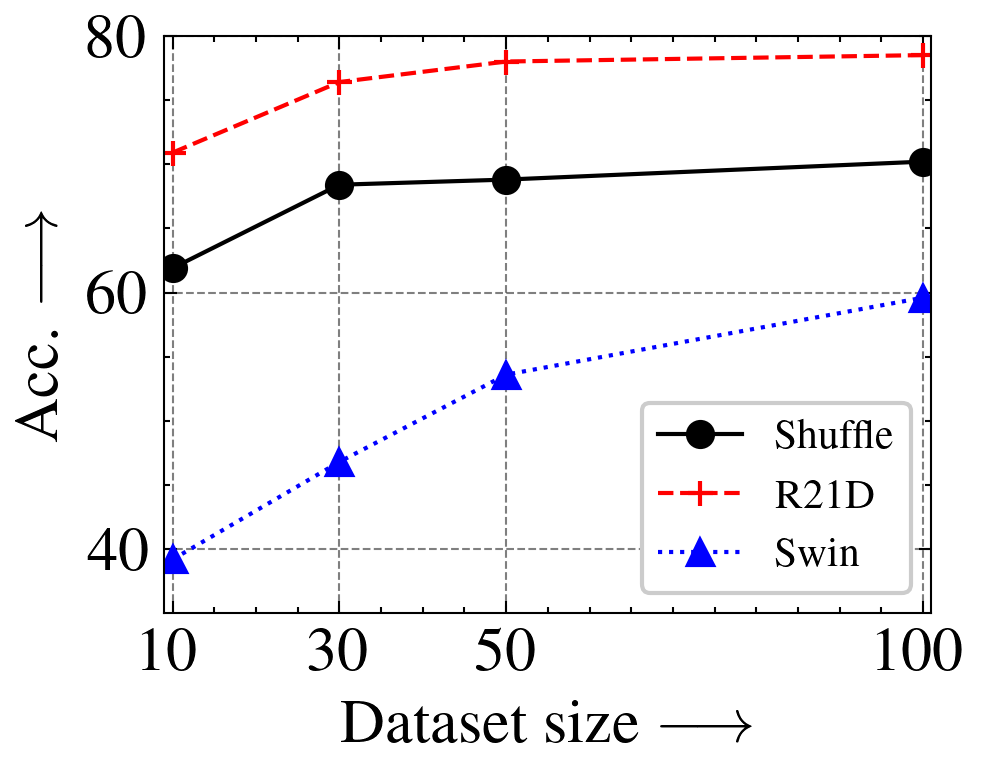}
    \caption{\textbf{Multiple architectures and data subsets on UCF101. } Pretext task is RSPNet. (x-axis: subset
size, y-axis: Top-1 Accuracy)  Here, 10 means 10k dataset subset, 30 means 30k and so on.}
    \label{fig:subset_analyze_ucf101}
\end{figure}

\begin{figure}[t!]
    \centering
      \includegraphics[width=0.8\linewidth]{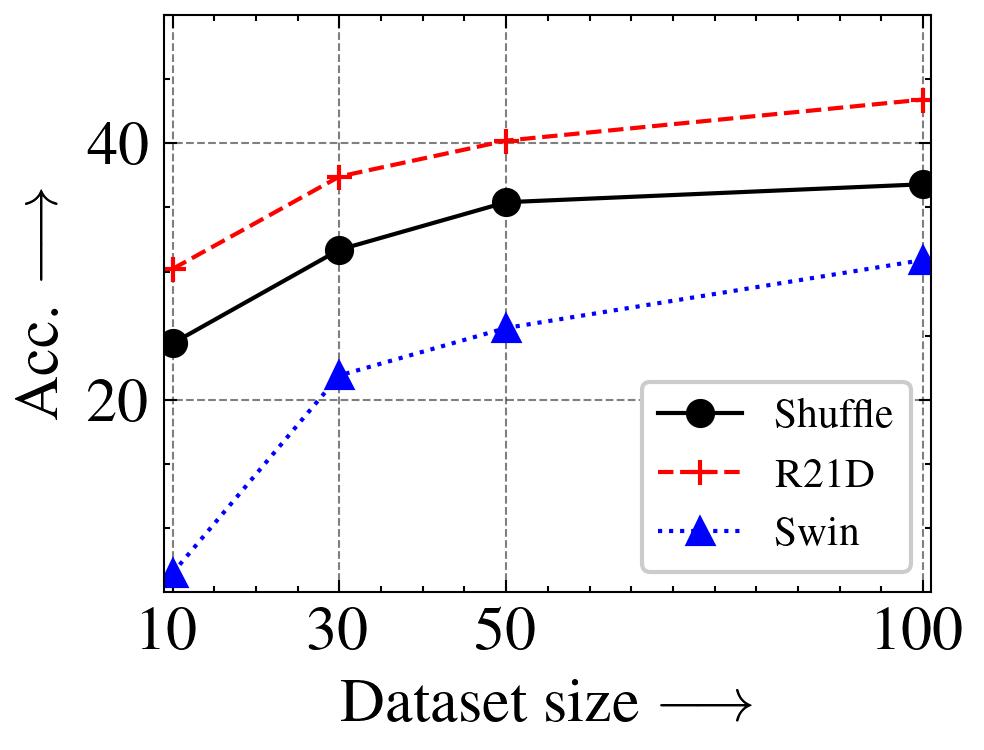}
    \caption{\textbf{Multiple architectures and data subsets on HMDB51. } Pretext task is RSPNet. (x-axis: subset
size, y-axis: Top-1 Accuracy)  Here, 10 means 10k dataset subset, 30 means 30k and so on.}
    \label{fig:subset_analyze_hmdb51}
\end{figure}

In Table \ref{tab:pretrain_subset_tasks_full}, we extend results for different pretext tasks on HMDB51 dataset. Similar to UCF101, \textit{the scale in subset size doesn't reciprocate to gain in performance} for all pretext tasks on HMDB51 dataset. From Figures \ref{fig:subset_analyze_ucf101} and \ref{fig:subset_analyze_hmdb51}, we see that performance increase for Swin by a good margin, whereas in case of ShuffleNet and R21D it's relatively less beyond 50k subset.
% \A{convert to plots and discuss it like std deviation plots}
\paragraph{Training time} Table \ref{tab:duration_rsp_ucf101} shows VideoSwin saturates at 150 epochs on UCF101 whereas CNN architectures saturates earlier (100 epochs) which reflects limitation of model capacity. Figure \ref{fig:cka_10k_epoch} shows the emergence of block structures for R21D network trained on RSPNet for K400 10k. The saturation point has reached earlier around 100 epochs which supports the hypothesis in main work that CNN architectures mostly saturates around 100 epochs. We see similar pattern even after increasing the dataset size.

 \begin{figure*}[t!]
     \centering
   \includegraphics[width =0.45\linewidth]{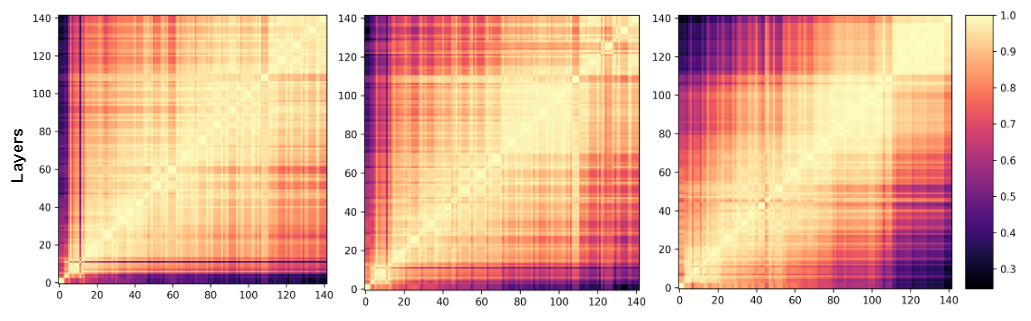} \hspace{1em}
   \includegraphics[width=0.45\linewidth]{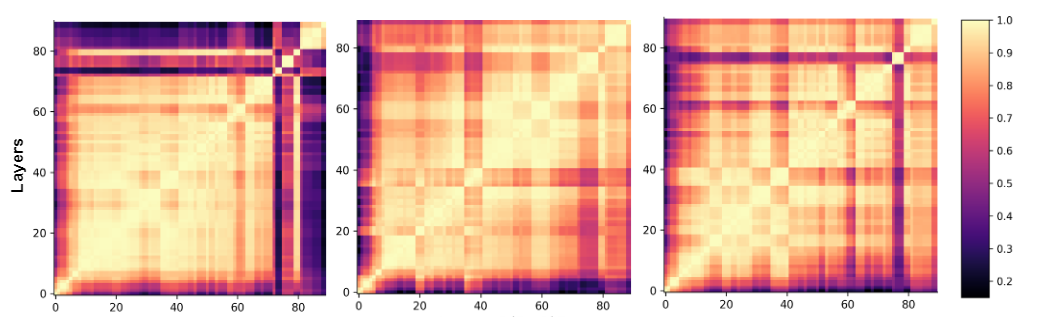}
     \caption{\textbf{Complexity CKA maps} PRP  ShuffleNet (Left) and R21D (Right) network increasing complexity from 2 to 4 (Left to right). ShuffleNet has lower performance than R21D, and it shows darkest patterns when complexity is increased from 3 to 4. For both of these complexities, R21D shows staggering grids.
     }
     % multi-block patterns for the complexities investigated, producing the darkest pattern 
\label{fig:cka_prpr21d}
 \end{figure*}

\begin{table*}[t!]
\centering
\begin{tabular}{c cccc cccc cccc}
% & \multicolumn{4}{c}{UCF101} \\
Epochs & \multicolumn{4}{c}{Shuffle} & \multicolumn{4}{c}{R21D} & \multicolumn{4}{c}{Swin}\\
\cmidrule(lr){2-5} \cmidrule(lr){6-9} \cmidrule(lr){10-13}
&10k & 30k & 50k & 100k & 10k & 30k & 50k & 100k & 10k & 30k & 50k & 100k \\
\hline
% shuffle/r21d/trans
50  & 59.1 & 66.3 & 68.1 & 68.9 & 66.8 & 71.1 & 75.0 & 77.2 & - & 40.4 & 44.9 & 52.0 \\
100  & 60.3 & 67.6 & 68.7 & 69.0 & 69.5 & 75.2 & 76.1 & 80.0 & 37.2 & 44.3 & 49.6 & 58.5\\
150  & 61.8 &  66.7 & 69.4 & 69.7 & 69.5 & 76.6 & 76.5 & 78.8 & 37.9 & 46.2 & 50.7 & 61.3\\
200  & 61.5 &  68.2 & 68.5 & 69.9 & 69.6 & 76.6 & 77.4 & 78.3 & 36.8 & 46.3 & 52.5 & 61.5\\
% transformer - 
\end{tabular}
\caption{RSPNet with different subset size on ShuffleNet/R21D/VideoSwin on UCF101 dataset.}
\label{tab:duration_rsp_ucf101}
\end{table*}

\subsection{Impact of task complexity}

Figures \ref{fig:cka_prpr21d} shows for ShuffleNet dark patterns with increase in complexity. R21D shows staggering grids. It supports our hypothesis that \textit{model capacity} plays an important role to learn meaningful features and always increasing the complexity doesn't reciprocate to \textit{better spatio-temporal features}. 

\begin{figure*}[t!]
     \centering
   \includegraphics[width =\linewidth]{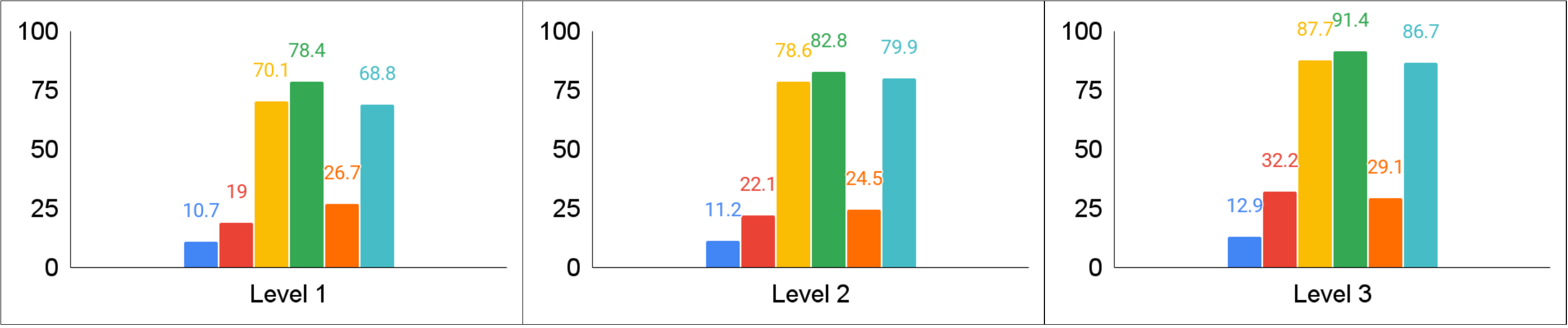}
     \caption{\textbf{Relative decrease in performance} at three different severity levels in increasing order from left to right. The pretext tasks is depicted by following colors - \textcolor{blue}{RotNet}, \textcolor{red}{VCOP}, \textcolor{yellow}{PRP}, \textcolor{green}{CVRL}, \textcolor{orange}{TDL}, \textcolor{cyan}{RSPNet}.}
\label{fig:noise_chart}
 \end{figure*}

\begin{figure}[t!]
\centering
   \includegraphics[width =\linewidth]{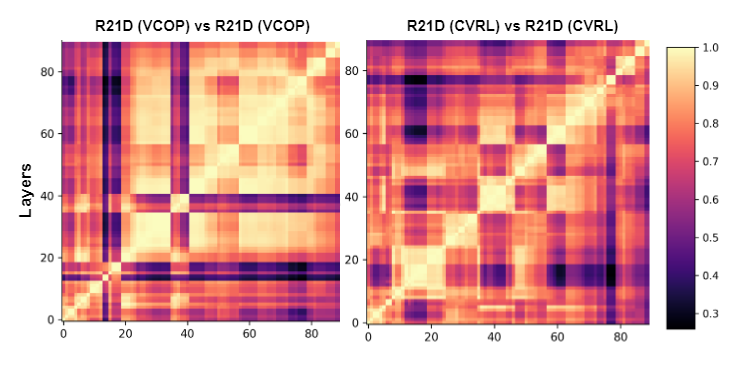}
     \caption{\textbf{Out-of-distribution CKA maps:} on VCOP and CVRL for R21D Network (Left to right). The semi-block structure of VCOP contrasts sharply with the grid-like structure of CVRL.}
\label{fig:cka_ood}
 \end{figure}

\subsection{Effect of data distrbituion}

\begin{table}[t!]
    \centering
    \footnotesize
    \begin{tabular}{l ccc ccc}
    \toprule
    & \multicolumn{3}{c}{Non-contrastive} & \multicolumn{3}{c}{Contrastive} \\
        \cmidrule(lr){2-4} \cmidrule(lr){5-7}
    & RotNet&VCOP&PRP& CVRL & TDL &RSP \\
    \midrule
    No Noise & 41.2 & 51.5 &46.2 &61.2 &31.7 & 78.0\\
    Gaussian &  40.9&47.0&14.6 &12.7& 28.0&16.7\\
    Impulse & 38.1&30.5&5.4 & 3.5& 18.8& 8.5\\
    Shot  &33.4&45.1& 20.9 &26.4 & 21.5&45.1\\
    Speckle & 34.7&43.9&14.4 &13.1 &24.7 & 27.0\\
    \bottomrule
    \end{tabular}
    \caption{Analysis of all pretext tasks with noise severity level 1 on R21D network on UCF101 dataset.}
    \label{tab:all_perturb}
\end{table}

Figure \ref{fig:cka_ood} illustrates CKA maps for networks pretrained on \textit{different source datasets} - for R21D pretrained on K400-50k on VCOP and CVRL respectively. The stark difference in semi-block structure of \textit{spatial (VCOP)} vs grid-like structure of \textit{spatio-temporal (CVRL)} shows spatio-temporal outperforms spatial pretext task.

\subsection{Robustness of SSL tasks}

Table \ref{tab:all_perturb} shows performance of each pretext on each type of noise for severity level 1. Fig. \ref{fig:noise_chart} shows a relative decrease in performance for three different severity level on UCF101 dataset. \textit{Non-contrastive} tasks are more robust than \textit{contrastive} on average even at different severity levels.

\subsection{Feature Analysis}
We employ knowledge distillation to evaluate how complementary information from different datasets can be used to train a student model that could take advantage of this in terms of performance gain and training time reduction. Here we show the numbers quantitatively. Table \ref{tab:kd_complexity} shows smaller architecture leans complementary information whereas bigger architecture depends on pretext task.  Table \ref{tab:kd_ood_ucf101} shows that for each pretext task, we learn \textit{complimentary information} from two \textit{different source} datasets. Thus, student always outperforms the teachers. Table \ref{tab:kd_categories} shows that distilling knowledge from a \textit{spatial} and a \textit{temporal} task outperforms the standalone \textit{spatio-temporal} task in both \textit{contrastive} and \textit{non-contrastive} case. 

\begin{table}[t!]
\centering

\begin{tabular}{c ccc}
\toprule
TC$\downarrow$& RotNet & VCOP & PRP \\
\midrule
T1 & 20.1/48.3 & 41.6/\textbf{56.8} & 24.2/38.9\\
T2 & 20.2/\textbf{58.3} & 41.8/54.8 & 18.1/44.4\\
T3 & 16.6/41.2 & 40.6/55.6 & 21.9/46.2\\
\midrule
S & \textbf{75.0}/56.6 & \textbf{75.4}/43.5 & \textbf{76.1}/\textbf{61.0} \\
\bottomrule
\end{tabular}
\caption{\textbf{Complexity variation} with at three levels as teachers (T1, T2, T3) for all three pretext tasks. TC: Task complexity. Results are shown on UCF101 with ShuffleNet/R21D as backbones. }
\label{tab:kd_complexity}
\end{table}

\begin{table}[t!]
\centering
\begin{tabular}{l c c c}
\toprule
& K400 (T1) & SSV2(T2) &  Student \\
\midrule
 RotNet & 36.2  & 42.5 & 59.8   \\
 VCOP &  50.4& 59.7&  67.6  \\
 CVRL & 56.9 & 34.7 & 66.6\\
 RSPNet & 76.4 & 69.5 & 80.2\\
 \bottomrule
\end{tabular}
\caption{\textbf{Out-of-Distribution} settings on UCF101 dataset using R21D network with teachers as different \textit{source} datasets.}
\label{tab:kd_ood_ucf101}
\end{table}

\begin{table}[t!]
\centering
\begin{tabular}{l c c c}
\toprule
& S (T1) & T(T2) &  Student   \\
\midrule
 Non-Contrastive & RotNet&VCOP  & 61.1  \\
 Contrastive & CVRL &TDL & 70.3    \\
% Spatial & RotNet, CVRL & 62.2  & \textbf{83.4}  \\
% Spatio-Temporal&  PRP, RSPNet & 63.6  & 65.9  \\
% All & & 62.8  & 64.1  \\
\bottomrule
\end{tabular}
\caption{\textbf{Knowledge distillation across different pretext tasks.} Teachers: ShuffleNet; Student: ShuffleNet.}
\label{tab:kd_categories}
\end{table}

\subsection{Clip retrieval}
In Table \ref{tab:rsp_cr}, we show clip retrieval across different architectures on HMDB51 and UCF101 dataset. Amongst small capacity networks, ShuffleNet outperforms others and in medium-capacity R21D outperforms.

\begin{table}[t!]
    \centering
    % \begin{subtable}[h]{0.5\textwidth}
    %     \centering
        
        \begin{tabular}{l  c c}
        \toprule
          Network  & Top@1 & Top@5 \\
        \midrule
        Squeeze &15.9/38.5&37.6/56.5\\
         Mobile &16.2/37.4&36.5/55.6\\
         Shuffle   & 19.3/43.1 & 42.0/62.1 \\
         \midrule
         C3D &19.9/43.2&43.4/61.6\\
         R3D &19.3/40.4& 42.5/60.2\\
         R21D & 18.2/42.7 & 40.1/62.8\\
         \bottomrule
        \end{tabular}
    \caption{Top K Clip Retrieval on HMDB51/UCF101 across different architectures for RSPNet.}
    \label{tab:rsp_cr}
\end{table}

\begin{table*}[t!]
\centering
\small
\begin{tabular}{l cccccc}
\toprule
Approach&Venue & NxW/H & Backbone &  Pre-training & UCF101 & HMDB51 \\
\midrule
\textbf{Generative}\\
\midrule
VIMPAC \cite{Tan2021VIMPACVP}&  - & 10x256 & ViT-L & HTM & 92.7 & 65.9 \\
VideoMAE \cite{videomae} & NeurIPS'22 & 16x224& ViT-B & K400& 91.3 & 62.6\\
VideoMAE $^{*}$ \cite{videomae} & NeurIPS'22 & 16x112& R21D-18 & K400& 76.2 & 45.4\\
\midrule
\textbf{Context} \\
\midrule
% V
PacePred \cite{pace} &  ECCV'20 &  16x112& R21D-18 & K400& 77.1 & 36.6\\
TempTrans \cite{temporals} & ECCV'20 & 16x112 & R3D-18& K400 & 79.3& 49.8 \\
STS \cite{stats} &TPAMI-21 & 16x112& R21D-18 & K400& 77.8 & 40.5\\
VideoMoCo \cite{videomoco} & CVPR'21& 16x112& R21D-18 & K400& 78.7 & 49.2\\
RSPNet \cite{rspnet} & AAAI'21& 16x112& R21D-18 & K400& 81.1 & 44.6\\
TaCo \cite{taco} & - & 16x224& R21D-18 & K400& 81.8 & 46.0\\
TCLR\cite{tclr} & CVIU'22& 16x112 & R21D-18 & K400& 88.2 & 60.0\\
CVRL$^{\dagger}$ \cite{cvrl} & CVPR'21& 32x224& R21D-18 & K400& 92.9 &67.9\\
TransRank \cite{transrank}&  CVPR'22 & 16x112 & R21D-18& K200& 87.8 & 60.1 \\

\midrule
\textbf{Multi-Modal}\\
\midrule
AVTS \cite{avts}& NeurIPS'18 &  25x224 & I3D & K400& 83.7 & 53.0 \\
GDT \cite{gdt} & - & 32x112& R21D & IG65M & 95.2 & 72.8\\
XDC \cite{xdc} & NeurIPS'20& 32x224& R21D &K400 & 84.2 & 47.1\\
\midrule
Ours $^{*}$ & - &16x112& R21D-18 &K400-30k & 97.3 & 51.5\\
\bottomrule
\end{tabular}
\caption{\textbf{Comparison with previous approaches} pre-trained on K400. Ours ( $^{*}$ best performing) is RSPNet pretrained on 30k subset of K400. $^{\dagger}$ modified backbone.}
\label{tab:action_recog_acc_full}
\end{table*}

\section{Main Table}
\label{sec:maintable}

In this section, we firstly expand the Table 6 (main paper) including results on HMDB51 dataset (Table \ref{tab:action_recog_acc_full}). Knowledge distilled network discussed in the main paper still shows competitive performance on HMDB51. Going in depth, the works outperforming us are AVTS\cite{avts}, GDT \cite{gdt} in multi-modal and VIMPAC \cite{Tan2021VIMPACVP}, VideoMAE \cite{videomae}, TCLR \cite{tclr} and CVRL \cite{cvrl} in single modality. AVTS and GDT uses two modalities, have more number of frames and AVTS also uses a bigger spatial size. Coming to Generative-based, both VIMPAC and VideoMAE uses a bigger backbone architecture.  CVRL uses a longer temporal sequence and bigger frame resolution compared to ours and TCLR utilize 64 effective frames. Thus, the performance on HMDB51 is still competitive.

\end{document}